\documentclass[11pt]{article}

\usepackage[final]{acl}

\usepackage{algorithm}
\usepackage{algorithmic}
\usepackage{url}
\usepackage{enumitem}

\usepackage{booktabs}
\usepackage{multirow}
\usepackage[utf8]{inputenc}
\usepackage{url}
\usepackage{booktabs}
\usepackage{amsmath}
\usepackage{amssymb}
\usepackage{nicefrac}
\usepackage{microtype}
\usepackage{algorithm}
\usepackage{algorithmic}
\usepackage{multirow}
\usepackage{subcaption}
\usepackage[table]{xcolor} 
\usepackage{enumitem}
\usepackage{booktabs}
\usepackage{amssymb}
\usepackage{amsthm}
\usepackage{mdframed}
\usepackage{booktabs} 
\usepackage{tcolorbox} 
\tcbuselibrary{skins,breakable}
\usepackage{booktabs}           
\usepackage{array}  
\usepackage{listings}
\DeclareCaptionStyle{ruled}{labelfont=normalfont,labelsep=colon,strut=off} 
\floatstyle{ruled}
\newfloat{listing}{tb}{lst}{}
\floatname{listing}{Listing}

\usepackage{times}
\usepackage{latexsym}

\usepackage[T1]{fontenc}

\usepackage[utf8]{inputenc}

\usepackage{microtype}

\usepackage{inconsolata}

\usepackage{graphicx}

\title{\textsc{Emotion2Skill}: Model-Internal Emotion Signals\\for Adaptive Skill Selection and Evolution}

\author{
\textbf{Bohan Lin\textsuperscript{1,2}},
\textbf{Hejia Geng\textsuperscript{3}},
\textbf{Xinyi Xie\textsuperscript{4}},
\textbf{Heng Zhou\textsuperscript{1,5}},
\\
\textbf{Qinghua Xing\textsuperscript{1}},
\textbf{Chen Zhang\textsuperscript{5}\thanks{Chen Zhang and Yudong Zhang are the corresponding authors.}},
\textbf{Yudong Zhang\textsuperscript{1,2,*}}
\\
\\
\textsuperscript{1}University of Science and Technology of China (USTC),\\
\textsuperscript{2}Suzhou Institute for Advanced Research, USTC,\\
\textsuperscript{3}University of Oxford,
\textsuperscript{4}University of Arizona,
\textsuperscript{5}Shanghai AI Laboratory
\\
\small{
\textbf{Contact:} \href{mailto:linbohan@ustc.edu}{linbohan@ustc.edu}
}
}

\begin{document}
\maketitle
\begin{abstract}
Skill-based LLM agents select reusable procedures from external libraries
to solve complex tasks, yet their routing decisions rely primarily on
task descriptions, verbal reflections, and experience-derived rules,
without explicitly observing the model's internal decision state.
Recent interpretability work has identified emotion representations
that influence model behavior, but their use has largely focused on
post-hoc analysis and output steering. Their potential to guide
agent-level decisions remains underexplored.
We propose \textsc{Emotion2Skill}, a framework that extracts LLM-internal
emotion vectors and incorporates them into both skill selection and
skill evolution. The framework uses these representations to inform
skill choice and tracks their changes to identify problematic
invocations, enabling targeted skill refinement beyond coarse outcome
feedback. Across two backbones, \textsc{Emotion2Skill} outperforms seven
baselines. On Qwen3-8B, it exceeds Trace2Skill by 3.1 percentage points
in WebShop success and 5.7 in ALFWorld average success.
Ablations and specificity controls support the utility of context-aligned
emotion signals, while co-activation analysis reveals semantically
coherent emotion--skill associations. The same emotion vectors also
transfer to MATH and MBPP without re-extraction.
These results establish LLM-internal emotion representations as an
effective decision-level signal for orchestrating agent skill systems,
extending their utility beyond interpretability and output steering.
Code is available at \href{https://github.com/BoHan-LIN04/Emotion2Skill}{Emotion2Skill}.

\end{abstract}

\section{Introduction}
\label{sec:intro}

Skill-based LLM agents solve complex, long-horizon interactive tasks
by retrieving reusable procedural knowledge from external
libraries~\cite{wang2023voyager,yang2026skillopt,yu2026masa,ni2026trace2skill}.
While recent work has substantially advanced skill
evolution~\cite{yang2026skillopt,yu2026masa}, trajectory
distillation~\cite{ni2026trace2skill}, and unified skill
training~\cite{li2026skill1}, the quality of skill \emph{selection}
continues to bound overall task performance.
No amount of skill refinement compensates for systematically retrieving
an inappropriate procedure at decision time.
Effective skill selection requires conditioning on two complementary
sources of information.
The first is the \emph{external textual context}: the task description,
conversation history, and environment observations that standard agents
already exploit.
The second is the model's own \emph{internal state}: its accumulated
uncertainty, its assessment of prior action outcomes, and the
trajectory-level confidence it has built up to the current step.
Identical external contexts can demand different skills depending on
this internal state, yet no existing skill-selection method directly
observes or leverages it.
\begin{figure}[t]
\centering
\includegraphics[width=\columnwidth]{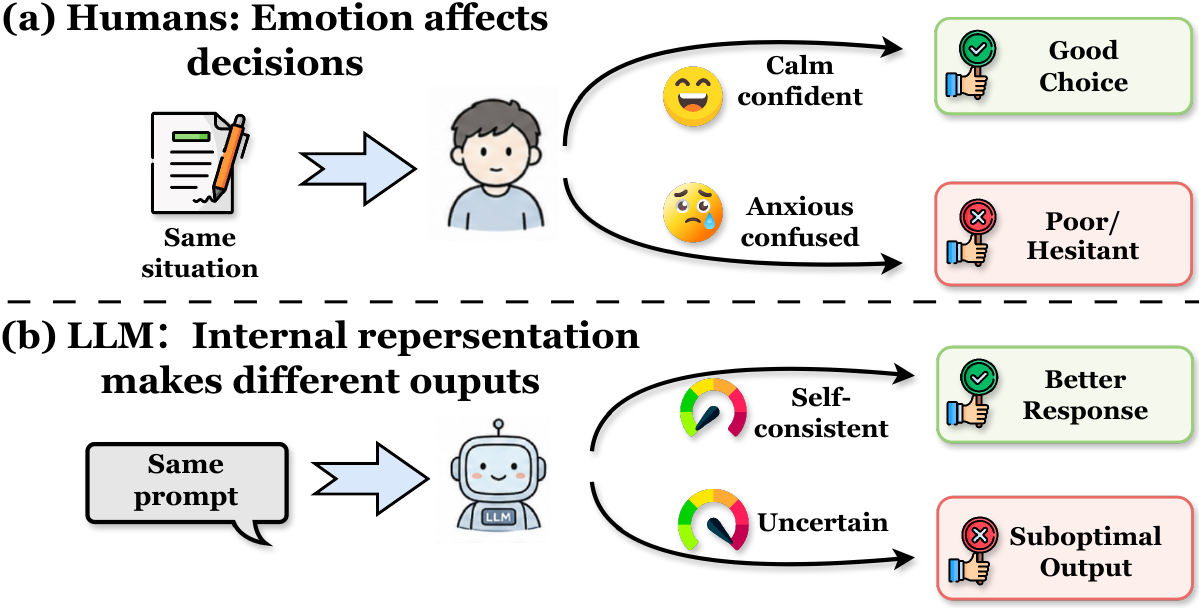}
\caption{Human decision-making integrates external stimuli with internal
affective states. We ask whether LLM agents can exploit an analogous
signal: the emotion representations encoded in the model's own residual
stream, extracted at inference time and used to condition skill utilization.}
\label{fig:motivation}
\end{figure}

Existing approaches to skill selection rely primarily on explicit
textual artifacts.
Text-embedding retrieval~\cite{wang2023voyager,yu2026masa} matches task
descriptions against skill descriptions in a shared semantic space,
verbal self-reflection~\cite{shinn2023reflexion} produces textual
critiques of prior actions to inform subsequent choices, experience-based
learning~\cite{zhao2024expel} distills trajectory-level lessons into
natural-language rules, and model-aware
alignment~\cite{yu2026masa} refines skill content through text-space
search.
Although effective and complementary, these methods share a structural
limitation, namely that they operate exclusively on the external textual
context while the model's internal representational state at the moment
of selection remains unobserved.
The latent component of the selection problem is therefore
systematically absent from the routing decision.

Neuroscience and psychology have long established that human decisions
are shaped not only by external stimuli but also by internal affective
states~\cite{damasio1994descartes,lerner2015emotion} (Figure~\ref{fig:motivation}).
We ask whether an analogous principle holds for LLMs.
Recent interpretability work shows that LLMs encode linear, causal
emotion representations in their residual
streams~\cite{zou2023representation,sofroniew2026emotions} that can be
read out at inference time and that \emph{causally} influence
agentically relevant outputs such as reward-hacking and sycophancy
rates~\cite{sofroniew2026emotions}.
Unlike generic hidden-state features, emotion vectors are empirically
linked to model behavior, interpretable by construction (each
dimension names a specific affect), and task-independent, making them
a natural proxy for the agent's internal decision state. \textbf{Skill selection should therefore incorporate the agent's internal decision state alongside external textual context, and emotion vectors offer an interpretable proxy for doing so.}

Building on this insight, we propose \textsc{Emotion2Skill}, a framework
that augments LLM-based skill selection with an explicit
emotion-derived signal.
At each decision step, the agent's emotion state is extracted from the
model's residual stream and passed through a lightweight encoder that
produces a natural-language emotion template together with a confidence
score. This emotion summary is concatenated into the LLM's selection
prompt as auxiliary context, and a confidence-gating mechanism omits the
signal when it is uninformative. Beyond selection, emotion trajectories
recorded during episodes are analyzed for abrupt internal state shifts,
and the resulting diagnostics guide periodic skill evolution by
directing the rewriter's attention to problematic trajectory segments.
On WebShop~\cite{yao2022webshop} and ALFWorld~\cite{shridhar2021alfworld},
\textsc{Emotion2Skill} with Qwen3-8B~\cite{qwen3} improves over
Trace2Skill~\cite{yu2026masa}, the strongest baseline, by \textbf{+5.7~pp}
average success on ALFWorld and \textbf{+3.1~pp} success rate on
WebShop, outperforming all baselines in aggregate with consistent
gains on Qwen3-14B.

Our contributions are as follows.
\begin{itemize}[leftmargin=*,itemsep=2pt,topsep=2pt]
  \item We reveal the connection between LLM-internal emotion vectors
    and agent skill selection, showing empirically that emotion
    representations encode decision-relevant information absent from
    text-level features.
  \item We design \textsc{Emotion2Skill}, a framework that leverages
    emotion vectors for both skill selection and skill evolution in
    LLM-based agents.
  \item We conduct extensive experiments on two interactive benchmarks
    against five baselines, together with ablation studies and
    interpretability analyses that validate the effectiveness of the
    proposed approach.
\end{itemize}

\section{Related Work}
\label{sec:related}

\paragraph{LLM internal representations and emotion vectors.}
A growing body of work shows that LLM internal representations encode
high-level semantic concepts as linear directions in activation
space~\cite{zou2023representation,park2023linear,nanda2023emergent}
that can be read out at inference time~\cite{burns2022latent,li2024probing}
and causally steered to alter model
behavior~\cite{turner2023actadd,rimsky2024contrastive}.
\citet{sofroniew2026emotions} demonstrate that frontier LLMs maintain
fine-grained emotion directions aligned with the 27-category GoEmotions
taxonomy~\cite{demszky2020goemotions} that causally influence
agentically relevant behaviors; concurrent mechanistic studies further
show that such emotion structure non-monotonically shapes agent
reasoning and safety~\cite{sun2026esteer,shou2026mechanistic}.
We adopt these categorical directions rather than appraisal
vectors~\cite{scherer2001appraisal} or valence-arousal subspaces, as
they are the representation for which causal influence on agent
behavior has been directly demonstrated, and named affective categories
provide interpretable routing rationales without additional grounding.
\textsc{Emotion2Skill} is the first to use emotion vectors as an
active routing and evolution signal for agent skill systems, moving
beyond post-hoc analysis and direct output steering.

\paragraph{Skill-based LLM agents.}
Skill-based LLM agents maintain reusable procedural knowledge for
long-horizon
tasks~\cite{wang2023voyager,yang2026skillopt,yu2026masa,ni2026trace2skill},
with recent work advancing skill evolution~\cite{yang2026skillopt,yu2026masa,yang2026skillmaster},
including EvoSkill~\cite{wang2026alignevoskill}, trajectory
distillation~\cite{ni2026trace2skill}, unified
training~\cite{li2026skill1}, and experiential
learning~\cite{zhao2024expel,shinn2023reflexion,xu2026ael}.
Concurrent work advances complementary dimensions of skill learning:
ARISE~\cite{li2026arise} introduces hierarchical reinforcement learning over
intrinsic skill representations; OPID~\cite{yang2026opid} applies on-policy
self-distillation for dense skill-level supervision; Skill-R1~\cite{vishe2026skillr1}
and ReSkill~\cite{he2026reskill} evolve skills via outcome-based and
RL-in-the-loop objectives; and MAGE~\cite{yang2026mage} co-evolves
skill knowledge graphs across multiple agents.
These approaches and \textsc{Emotion2Skill} operate on different
dimensions: where they optimize skill routing through external reward
or similarity feedback, \textsc{Emotion2Skill} conditions on the
model's internal affective state, and the two classes of signal are
complementary rather than directly comparable.

\section{Method}
\label{sec:method}

\begin{figure*}
    \centering
    \includegraphics[width=1\linewidth]{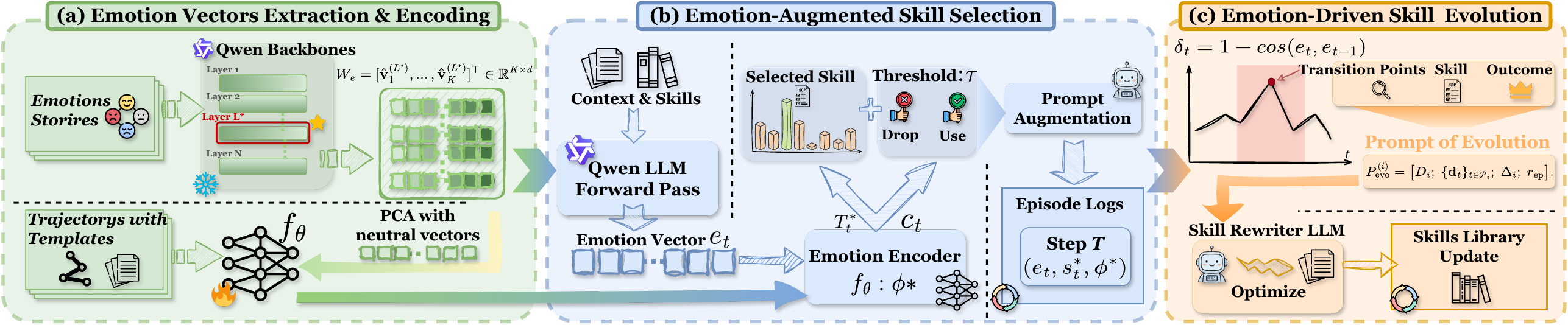}
    \caption{Framework of our method.
Left: offline extraction of emotion vectors from the LLM residual stream and MLP encoder training.
Middle: emotion state mapped to a confidence-gated summary injected into the skill-selection prompt.
Right: emotion trajectory shifts trigger targeted SOP rewriting for underperforming skills.}
    \label{fig:framework}
\end{figure*}

\textsc{Emotion2Skill} grounds skill decisions in the model's own affective
representations, filling the gap left by context-only skill routing.
\textbf{Emotion Vectors} are extracted from the residual stream and projected
into a natural-language summary at each step.
\textbf{Emotion-Augmented Skill Selection} injects this summary into the
routing prompt via a confidence gate.
\textbf{Emotion-Driven Skill Evolution} detects abrupt trajectory shifts
to localize and rewrite underperforming skills.
Figure~\ref{fig:framework} illustrates the design.

\subsection{Preliminaries}
\label{sec:formulation}

Consider an agent equipped with a skill library
$\mathcal{S} = \{s_1, \ldots, s_M\}$, where each skill $s_i$ is a
reusable procedure described by a natural-language SOP $D_i$.
At decision step $t$, the agent observes the external textual context
$x_t$ (task instruction, conversation history, and environment
observations) and possesses a latent internal decision state encoded
in the residual stream of the underlying LLM.
We denote by $\mathbf{h}_t^{(\ell)} \in \mathbb{R}^d$ the
residual-stream activation at layer $\ell$ and token position $t$,
and by $L^*$ the optimal extraction layer selected on the GoEmotions
validation split.

Standard skill-based agents select a skill $s_t^*$ by prompting the
LLM $\mathcal{M}$ with $x_t$ and the candidate SOPs
$\{D_1, \ldots, D_M\}$, relying entirely on the textual content of
these inputs.
\textsc{Emotion2Skill} extends this formulation by additionally
conditioning on an emotion summary $\boldsymbol{\phi}_t$ derived from
the agent's internal state:
\begin{equation}
s_t^* = \mathcal{M}\bigl(x_t,\;
  \{D_1, \ldots, D_M\},\;
  \boldsymbol{\phi}_t\bigr),
\label{eq:routing}
\end{equation}
where $\boldsymbol{\phi}_t$ is a natural-language token sequence
produced by a learned emotion encoder $f_\theta$.
The LLM remains the primary decision-maker; $\boldsymbol{\phi}_t$
serves as auxiliary context that makes the agent's latent state
explicit.


\subsection{Emotion Vector Extraction and Encoding}
\label{sec:extraction}

\paragraph{Contrastive extraction.}
We extract emotion directions from the agent model following the
contrastive-averaging procedure of
\citet{sofroniew2026emotions}, replicated on Qwen3.
For each of the $K{=}27$ emotion concepts in the GoEmotions
taxonomy~\cite{demszky2020goemotions}, the model generates $N{=}100$
short stories with the target emotion.
A forward pass through each story yields activations
$\mathbf{h}_t^{(\ell)}$ at every layer $\ell$ and token position $t$.
We mean-pool over tokens past an initial prefix ($t \geq 50$) to
obtain a per-story representation $\bar{\mathbf{h}}^{(\ell)}$, then
compute the emotion direction for concept $k$ at layer $\ell$ by
subtracting the global mean from the within-concept mean:
\begin{equation}
\mathbf{v}_k^{(\ell)} =
  \frac{1}{N}\sum_{n=1}^{N}\bar{\mathbf{h}}^{(\ell)}_{k,n}
  \;-\;
  \frac{1}{KN}\sum_{k'=1}^{K}\sum_{n=1}^{N}\bar{\mathbf{h}}^{(\ell)}_{k',n}.
\label{eq:contrastive}
\end{equation}
To suppress variance attributable to stylistic rather than affective
content, a PCA-based denoising step projects out the top principal
components of emotionally neutral activations, yielding denoised directions
$\hat{\mathbf{v}}_k^{(\ell)} = \Pi^\perp \mathbf{v}_k^{(\ell)}$.
The optimal layer $L^*$ is selected from a candidate grid by
maximizing GoEmotions classification accuracy 
.
Stacking the denoised directions yields the emotion extractor matrix
$W_e = [\hat{\mathbf{v}}_1^{(L^*)}, \ldots,
\hat{\mathbf{v}}_{K}^{(L^*)}]^\top \in \mathbb{R}^{K \times d}$.

At inference time, the raw emotion state at decision step $t$ is
obtained by projecting the residual-stream activation at the
\texttt{Assistant:} delimiter token~\cite{sofroniew2026emotions}
through $W_e$ and applying $\ell_2$ normalization:
\begin{equation}
\mathbf{e}_t = \frac{W_e\,\mathbf{h}_T^{(L^*)}}
  {\lVert W_e\,\mathbf{h}_T^{(L^*)} \rVert_2}
  \;\in \mathbb{R}^{K}.
\label{eq:emotionstate}
\end{equation}
The resulting vector $\mathbf{e}_t$ assigns a scalar activation to each
of the $K$ emotion dimensions, capturing the agent's affective state at
the moment immediately preceding generation.

\paragraph{Emotion encoder.}
The raw vector $\mathbf{e}_t$ is not directly consumable by the LLM's
text-based prompt; naively reporting the top-$k$ activated emotion
names as tokens discards magnitude information and ignores interactions
between co-activated emotions.
We therefore introduce a lightweight \emph{emotion encoder}
$f_\theta$ that maps $\mathbf{e}_t$ to two outputs used for prompt
augmentation.
The encoder is a 3-layer MLP
that produces an encoded representation
$\mathbf{r}_t = f_\theta(\mathbf{e}_t) \in \mathbb{R}^{d'}$.
From $\mathbf{r}_t$ we derive (i)~a \emph{template index} over a
predefined bank $\mathcal{T} = \{T_1, \ldots, T_C\}$ of $C{=}12$
natural-language emotion-state descriptions, where each template $T_j$
is associated with a learnable prototype $\mathbf{q}_j \in
\mathbb{R}^{d'}$:
\begin{equation}
T_t^* = \arg\max_{T_j \in \mathcal{T}}\;
  \frac{\mathbf{r}_t^\top \mathbf{q}_j}
  {\lVert\mathbf{r}_t\rVert\;\lVert\mathbf{q}_j\rVert},
\label{eq:template}
\end{equation}
and (ii)~a scalar \emph{confidence score}
\begin{equation}
c_t = \sigma(\mathbf{w}_c^\top \mathbf{r}_t + b_c) \in [0,1]
\label{eq:confidence}
\end{equation}
indicating how informative the current emotion state is for routing.
We define the composite emotion summary as the pair
$\boldsymbol{\phi}_t = (T_t^*,\, c_t)$.
When $c_t < \tau$, $\boldsymbol{\phi}_t$ is omitted from the prompt
entirely (confidence gating), allowing the system to fall back to
standard text-only selection.

The encoder is trained on tuples
$(\mathbf{e}_t, s_t^{\text{sel}}, y_t)$ collected during a warm-up
phase in which the agent runs episodes under a baseline policy
(ReAct with the skill library, without emotion signals), where
$s_t^{\text{sel}} \in \mathcal{S}$ is the baseline's skill choice and
$y_t \in \{0,1\}$ is the episode outcome.
The objective combines a supervised-contrastive
loss~\cite{khosla2020supcon} that clusters representations by
effective skill assignments with a template-classification loss.
For an anchor $\mathbf{r}_i$ with positive set
$\mathcal{A}^+(i) = \{j \mid s_j^{\text{sel}} {=} s_i^{\text{sel}}
\wedge y_j {=} 1\}$, the contrastive term is
\begin{equation}
\resizebox{0.95\linewidth}{!}{
$
\mathcal{L}_{\text{ctr}} = -\frac{1}{|\mathcal{B}|}
  \sum_{i \in \mathcal{B}}
  \frac{1}{|\mathcal{A}^+(i)|}
  \sum_{j \in \mathcal{A}^+(i)}
  \log \frac{
    \exp(\mathbf{r}_i^\top \mathbf{r}_j / \kappa)
  }{
    \sum_{k \neq i} \exp(\mathbf{r}_i^\top \mathbf{r}_k / \kappa)
  },
$
}
\label{eq:contrastive_loss}
\end{equation}
where $\kappa$ is a temperature and representations are
$\ell_2$-normalized.
The classification term assigns pseudo-labels $\hat{c}_t$ via
$k$-means ($k{=}C$) on the raw emotion vectors and minimizes the
cross-entropy over prototype similarities:
\begin{equation}
\mathcal{L}_{\text{cls}} = -\frac{1}{|\mathcal{B}|}
  \sum_{i \in \mathcal{B}}
  \log \frac{
    \exp(\mathbf{r}_i^\top \mathbf{q}_{\hat{c}_i} / \kappa')
  }{
    \sum_{j=1}^{C}
    \exp(\mathbf{r}_i^\top \mathbf{q}_{j} / \kappa')
  }.
\label{eq:cls_loss}
\end{equation}
The overall objective is
$\mathcal{L} = \mathcal{L}_{\text{ctr}} + \lambda\,
\mathcal{L}_{\text{cls}}$,
where $\lambda$ balances the two terms.


\subsection{Emotion-Augmented Skill Selection}
\label{sec:selection}

The emotion summary $\boldsymbol{\phi}_t$ produced by the encoder is
concatenated into the LLM's input prompt as additional context,
making the agent's latent state visible alongside the task
description and candidate skill SOPs.
At each decision step $t$, the system constructs an augmented prompt
$P_t$ by appending the emotion template $T_t^*$ and the confidence
score $c_t$ to the standard prompt components.
When the encoder's confidence falls below the threshold $\tau$, the
emotion fields are withheld, and the prompt reduces to the
conventional text-only format:
\begin{equation}
P_t =
\begin{cases}
  \bigl[x_t;\; \{D_i\}_{i=1}^M;\; T_t^*;\; c_t\bigr]
    & \text{if } c_t \geq \tau, \\[4pt]
  \bigl[x_t;\; \{D_i\}_{i=1}^M\bigr]
    & \text{otherwise}.
\end{cases}
\label{eq:prompt}
\end{equation}
The LLM then reads $P_t$ in its entirety and produces a skill
decision via constrained generation over the valid skill identifiers:
\begin{equation}
s_t^* = \arg\max_{s_i \in \mathcal{S}}\;
  p_{\mathcal{M}}(s_i \mid P_t),
\label{eq:llm_select}
\end{equation}
where $p_{\mathcal{M}}(s_i \mid P_t)$ denotes the conditional
probability that $\mathcal{M}$ assigns to skill $s_i$.
The emotion summary enters through the same input channel as
the task description, and setting $\tau{=}1$ recovers standard
text-only selection.


\subsection{Emotion-Driven Skill Evolution}
\label{sec:evolution}

Beyond online selection, the episode log $\mathcal{L}_{\text{ep}}$
provides a temporally ordered trace of internal states that can
inform periodic skill evolution.
Standard evolution methods~\cite{yang2026skillopt,yu2026masa} prompt
the LLM to rewrite underperforming skill SOPs based on binary
episode outcomes and, optionally, textual self-reflection.
This signal is coarse in that it arrives only after episode
termination and provides no indication of \emph{which} intermediate
steps were problematic.

\textsc{Emotion2Skill} augments the rewriter's input with structured
emotion diagnostics that localize critical state changes within the
episode.

Let $\mathcal{E} = (\mathbf{e}_1, \ldots, \mathbf{e}_T)$ denote the
emotion trajectory of a completed episode.
To identify steps at which the agent's internal state shifted
abruptly, we compute the step-to-step \emph{emotion shift magnitude}
\begin{equation}
\delta_t = 1 - \frac{\mathbf{e}_t^\top \mathbf{e}_{t-1}}
  {\lVert\mathbf{e}_t\rVert\;\lVert\mathbf{e}_{t-1}\rVert},
  \quad t = 2, \ldots, T.
\label{eq:shift}
\end{equation}
This quantity measures the cosine distance between consecutive
emotion states and is agnostic to which specific dimensions change;
large $\delta_t$ indicates a sharp reorganization of the agent's
internal state regardless of whether the shift involves
positive-valence, negative-valence, or cognitive dimensions.
An \emph{emotion transition point} is defined as a step at which
$\delta_t$ exceeds the episode mean by more than one standard
deviation:
\begin{equation}
\mathcal{P} = \bigl\{t \in \{2,\ldots,T\} \;\big|\;
  \delta_t > \bar{\delta} + \sigma_\delta\bigr\},
\label{eq:transition}
\end{equation}
where
$\bar{\delta} = (T{-}1)^{-1}\sum_{t=2}^{T} \delta_t$ and
$\sigma_\delta = \bigl((T{-}1)^{-1}\sum_{t=2}^{T}
(\delta_t - \bar{\delta})^2\bigr)^{1/2}$.
By operating on the full $K$-dimensional emotion vector without
pre-selecting specific dimensions, this criterion captures any
substantial internal reorganization, including shifts from
exploratory states to frustrated states, from confident states to
confused states, or from one cognitive mode to another.

For each skill $s_i$ flagged for evolution, we restrict to
skill-specific transitions
$\mathcal{P}_i = \{t \in \mathcal{P} \mid s_t^* = s_i\}$ and
extract a diagnostic context around each:
\begin{equation}
\mathbf{d}_t = (x_{t-1:t+1},\; s_{t-1:t+1}^*,\;
  T_{t-1:t+1}^*,\; \delta_{t-1:t+1}),
\label{eq:diagnostic}
\end{equation}
capturing a 3-step window of observations, skill choices, emotion
templates, and shift magnitudes.
These diagnostics, together with a verbalized summary $\Delta_i$
describing the transition pattern, are appended to the evolution
prompt alongside the current SOP $D_i$ and the episode outcome
$r_{\text{ep}}$:
\begin{equation}
P_{\text{evo}}^{(i)} = \bigl[
  D_i;\;
  \{\mathbf{d}_t\}_{t \in \mathcal{P}_i};\;
  \Delta_i;\;
  r_{\text{ep}}
\bigr].
\label{eq:evoprompt}
\end{equation}
The rewriter receives $P_{\text{evo}}^{(i)}$ as its full input
context and generates the revised SOP:
\begin{equation}
D_i' = \mathcal{M}\bigl(P_{\text{evo}}^{(i)}\bigr).
\label{eq:rewrite}
\end{equation}
Evolution proceeds over $R$ rounds; at each round the agent executes
a batch of episodes, identifies transition points, and updates each
underperforming skill (success rate $< \eta$):
\begin{equation}
\mathcal{S}^{(r+1)} = \bigl(\mathcal{S}^{(r)} \setminus
  \{s_i\}\bigr) \cup \{s_i'\},
  \quad \forall\, s_i \in \mathcal{U}^{(r)},
\label{eq:library_update}
\end{equation}
where $\mathcal{U}^{(r)}$ is the set of underperforming skills in
round $r$.
The emotion diagnostics enter through the same prompt channel as the SOP and
episode transcript, providing temporally localized context that
complements the coarse binary outcome signal.
Algorithm~\ref{alg:emotionskill} summarizes both the online selection
and the evolution procedure.

\begin{table*}[h!]
\centering\small
\setlength{\tabcolsep}{3.8pt}
\begin{tabular}{cl ccccccc cc}
\toprule
 & & \multicolumn{7}{c}{\textbf{ALFWorld (Success \%)}} & \multicolumn{2}{c}{\textbf{WebShop}} \\
\cmidrule(lr){3-9}\cmidrule(lr){10-11}
\textbf{Backbone} & \textbf{Method}
& \textbf{Pick} & \textbf{Look} & \textbf{Clean} & \textbf{Heat}
& \textbf{Cool} & \textbf{Pick2} & \textbf{Avg.}
& \textbf{Score} & \textbf{Succ.} \\
\midrule
\multirow{8}{*}{\rotatebox{90}{Qwen3-8B}}
 & Zero-Shot
   & 43.4 & 39.1 & 19.1 & 9.6 & 12.3 & 4.4
   & $21.9{\pm}1.5$ & 11.2 & $2.8{\pm}0.5$ \\
 & ReAct~\cite{yao2022react}
   & 53.1 & 42.0 & 41.3 & 20.7 & 18.4 & 3.6
   & $30.7{\pm}1.2$ & 12.1 & $5.2{\pm}3.2$ \\
 & Reflexion~\cite{shinn2023reflexion}
   & 52.2 & 34.4 & 29.0 & 24.1 & 19.1 & 14.1
   & $31.5{\pm}1.6$ & 22.4 & $6.8{\pm}2.4$ \\
 & ExpeL~\cite{zhao2024expel}
   & 61.2 & 30.1 & 44.4 & 35.6 & 28.7 & 11.5
   & $37.0{\pm}4.7$ & 39.6 & $18.8{\pm}4.1$ \\
 & MASA~\cite{yu2026masa}
   & 55.1 & 24.1 & 46.6 & 48.9 & \textbf{43.9} & 23.4
   & $43.3{\pm}0.9$ & 44.7 & $27.1{\pm}0.8$ \\
 & EvoSkill~\cite{wang2026alignevoskill}
   & 70.1 & 32.1 & 57.3 & 47.3 & 39.5 & 15.3
   & $46.4{\pm}1.9$ & 53.3 & $29.7{\pm}1.6$ \\
 & Trace2Skill~\cite{ni2026trace2skill}
   & \textbf{70.5} & 28.4 & 56.8 & 32.3 & 37.6 & 26.8
   & $47.4{\pm}0.9$ & 52.0 & $30.2{\pm}3.8$ \\
 & \textsc{Emotion2Skill} (Ours)
   & 61.2 & \textbf{62.5} & \textbf{63.3} & \textbf{57.8} & 42.7 & \textbf{35.3}
   & $\mathbf{53.1{\pm}1.6}$ & \textbf{64.7} & $\mathbf{33.3{\pm}0.6}$ \\
\midrule
\multirow{8}{*}{\rotatebox{90}{Qwen3-14B}}
 & Zero-Shot
   & 55.1 & 42.9 & 8.2 & 7.9 & 7.7 & 13.1
   & $23.4{\pm}2.8$ & 27.5 & $9.4{\pm}0.8$ \\
 & ReAct~\cite{yao2022react}
   & 52.1 & 36.6 & 19.8 & 19.0 & 16.5 & 17.8
   & $30.2{\pm}2.4$ & 37.6 & $13.0{\pm}1.6$ \\
 & Reflexion~\cite{shinn2023reflexion}
   & 59.0 & 33.8 & 31.6 & \textbf{46.9} & 10.2 & 13.8
   & $34.4{\pm}0.8$ & 32.7 & $14.4{\pm}7.0$ \\
 & ExpeL~\cite{zhao2024expel}
   & 71.0 & 35.2 & \textbf{43.8} & 37.0 & 34.3 & 28.9
   & $43.8{\pm}0.8$ & 45.4 & $20.6{\pm}1.2$ \\
 & MASA~\cite{yu2026masa}
   & \textbf{81.0} & 41.2 & 50.5 & 36.5 & \textbf{56.2} & 17.2
   & $49.2{\pm}0.8$ & 46.8 & $24.8{\pm}2.0$ \\
 & EvoSkill~\cite{wang2026alignevoskill}
   & 72.9 & 43.4 & 56.3 & 41.6 & 42.3 & 23.9
   & $49.7{\pm}1.9$ & 56.0 & $31.2{\pm}1.6$ \\
 & Trace2Skill~\cite{ni2026trace2skill}
   & 59.3 & \textbf{58.0} & 48.6 & 52.0 & 45.5 & \textbf{38.8}
   & $52.1{\pm}2.6$ & 50.7 & $32.8{\pm}1.7$ \\
 & \textsc{Emotion2Skill} (Ours)
   & 68.3 & 35.6 & \textbf{65.2} & \textbf{71.7} & \textbf{63.1} & 34.2
   & $\mathbf{57.0{\pm}1.6}$ & \textbf{66.8} & $\mathbf{37.0{\pm}0.6}$ \\
\bottomrule
\end{tabular}
\caption{Main results on ALFWorld and WebShop.
All values are means over three runs; ALFWorld Avg.\ and WebShop Succ.\
additionally report standard deviation.
\textbf{Bold} denotes the best result per column .
\textsc{Emotion2Skill} achieves the best aggregate performance on both
benchmarks across Qwen3-8B and Qwen3-14B.}
\label{tab:main_results}
\end{table*}

\section{Experiments}
\label{sec:experiments}

\subsection{Setup}
\label{sec:setup}

We evaluate on \textbf{WebShop}~\cite{yao2022webshop} for web-based
shopping and \textbf{ALFWorld}~\cite{shridhar2021alfworld} for
text-based household tasks. On WebShop, we report Score and success
rate. On ALFWorld, we report per-type and average success rates.
We compare against Zero-Shot, ReAct~\cite{yao2022react},
Reflexion~\cite{shinn2023reflexion},
ExpeL~\cite{zhao2024expel}, MASA~\cite{yu2026masa},
EvoSkill~\cite{wang2026alignevoskill},
and Trace2Skill~\cite{ni2026trace2skill};
all methods are reproduced on both Qwen3-8B and Qwen3-14B~\cite{qwen3}.
All experiments are conducted on 2$\times$H800 GPUs.
\textsc{Emotion2Skill} uses confidence threshold $\tau{=}0.3$, with the
extraction layer $L^*$ selected by hit-any classification accuracy on
the GoEmotions validation split.
Qwen3-14B reuses the same skill libraries and $\tau$; only $L^*$
is re-selected.
Dataset statistics, baseline reproduction protocols, hyperparameters,
prompt templates, and skill libraries are detailed in
Appendices~\ref{app:exp_setup} and~\ref{app:prompts_skills}.

\vspace{-2mm}
\subsection{Emotion Vector Quality and Layer Selection}
\label{sec:preliminary}

We test whether Qwen3-8B's residual stream encodes discriminable
emotion structure by measuring classification accuracy across the 27 emotion
categories in the GoEmotions validation split~\cite{demszky2020goemotions}.
We select the extraction layer~$L^*$ using this validation accuracy.

\begin{figure}[t]
\centering
\includegraphics[width=\columnwidth]{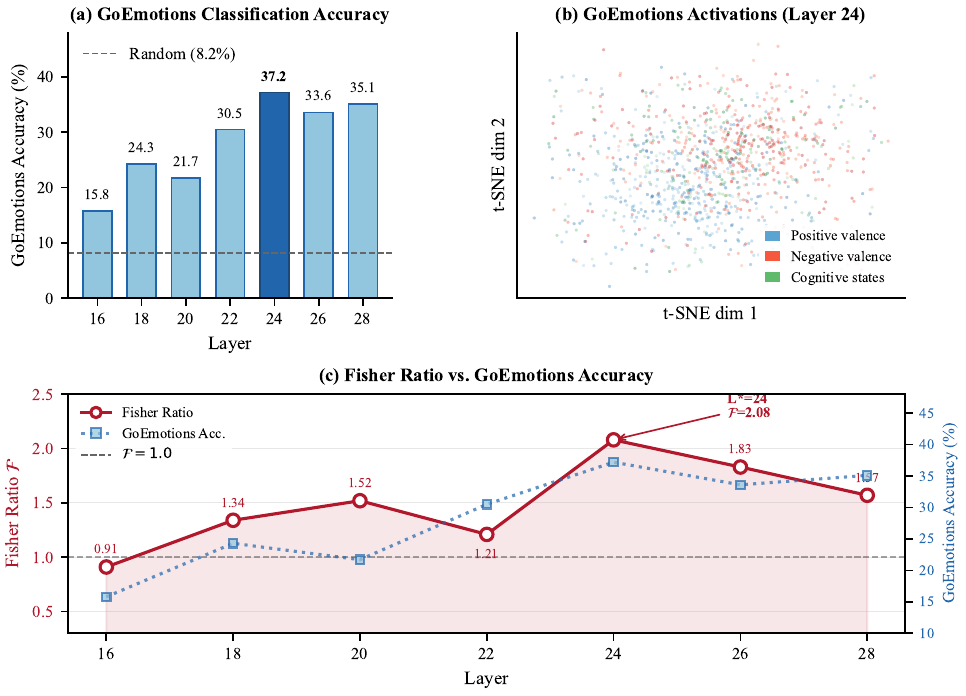}
\caption{Emotion vector quality and layer selection for Qwen3-8B.
(a)~GoEmotions validation accuracy across layers peaks at 37.2\% at
layer~24; the dashed line marks the 8.2\% random baseline.
(b)~t-SNE of held-out GoEmotions activations at $L^*{=}24$; valence
regions are distinguishable but show substantial overlap.
(c)~Fisher discriminability ratio peaks at $\mathcal{F}{=}2.08$ at
layer~24 but does not track accuracy monotonically.}
\label{fig:preliminary}
\end{figure}

\paragraph{Evaluation protocol and layer selection.}
For the 36-layer Qwen3-8B model, we evaluate candidate layers
$\ell \in \{16, 18, 20, 22, 24, 26, 28\}$. At each layer, we project
the GoEmotions validation activations onto the
contrastively-extracted directions $\hat{\mathbf{v}}_k^{(\ell)}$ and
classify each example by its highest-cosine-similarity direction,
without any task-specific fine-tuning.
We choose $L^*$ using only this validation accuracy; the GoEmotions
test split is held out from layer selection and is used only for
post-selection evaluation or visualization.
Since GoEmotions is multi-label, we adopt a \textit{hit-any} criterion:
a prediction is correct if the top direction matches any annotated
label. Under this criterion, a random baseline achieves 8.2\% accuracy.
Figure~\ref{fig:preliminary}a shows that accuracy is non-monotonic
across layers. It rises from 15.8\% at layer~16 to 24.3\% at layer~18,
dips to 21.7\% at layer~20, and then climbs through 30.5\% at layer~22
to a peak of $\mathbf{37.2\%}$ at \textbf{layer~24}. Accuracy falls to
33.6\% at layer~26 before recovering to 35.1\% at layer~28.
The peak at $24/36{=}0.67$ depth is broadly consistent with
\citeauthor{sofroniew2026emotions}'s~(\citeyear{sofroniew2026emotions})
observation that emotion representations concentrate near two-thirds
depth, though the non-monotonic profile suggests additional
layer-specific specialization.
We fix $L^*{=}24$ for all Qwen3-8B experiments. For the 40-layer
Qwen3-14B model, the same validation-only sweep over
$\{18, 20, 22, 24, 26, 28, 30\}$ selects $L^*{=}26$ with 39.4\%
accuracy.
\vspace{-5mm}
\paragraph{Geometric structure.}
Figure~\ref{fig:preliminary}b visualizes GoEmotions activations at
layer~24 via t-SNE. Positive-valence emotions such as \textit{joy},
\textit{love}, and \textit{optimism} concentrate in the lower-left
quadrant, while negative-valence emotions such as \textit{anger},
\textit{disgust}, and \textit{sadness} occupy the upper-right.
Cognitive states such as \textit{curiosity} and \textit{confusion}
form an intermediate region. Closely related affects within the same
valence band still overlap, as seen for \textit{anger} and
\textit{annoyance}.
Fisher's ratio
$\mathcal{F}{=}\sigma^2_{\text{between}}/\sigma^2_{\text{within}}$
peaks at $\mathcal{F}{=}\mathbf{2.08}$ at layer~24 but does not track
accuracy perfectly. Layer~20 has $\mathcal{F}{=}1.52$ and 21.7\%
accuracy, whereas layer~22 achieves 30.5\% accuracy despite a lower
ratio of $\mathcal{F}{=}1.21$. Thus, the ratio serves as a separability
diagnostic rather than a substitute for validation accuracy.

\subsection{Main Results}
\label{sec:main_results}

\paragraph{Overall comparison.}
Table~\ref{tab:main_results} reports results on both backbones.
\textsc{Emotion2Skill} attains the best aggregate score on both
benchmarks at both scales.
On Qwen3-8B it reaches \textbf{53.1\%} average success on ALFWorld and
\textbf{33.3\%} Succ.\ on WebShop, exceeding Trace2Skill~\cite{ni2026trace2skill},
the strongest aggregate baseline, by \textbf{+5.7} and \textbf{+3.1}~pp;
on Qwen3-14B it reaches \textbf{57.0\%} and \textbf{37.0\%}, a margin of
\textbf{+4.9} and \textbf{+4.2}~pp over the same baseline.
Its standard deviation is at most 1.6~pp across these settings.
By comparison, ExpeL has a standard deviation of 4.7~pp on ALFWorld
with Qwen3-8B, and Reflexion has a standard deviation of 7.0~pp on
WebShop with Qwen3-14B.

\paragraph{Per-task behavior and scaling.}
On Qwen3-8B, \textsc{Emotion2Skill} exceeds Trace2Skill by 34.1~pp on
Look, 25.5~pp on Heat, and 8.5~pp on Pick2. Trace2Skill remains ahead
by 9.3~pp on Pick. Relative to EvoSkill, the largest gains are
30.4~pp on Look, 20.0~pp on Pick2, and 10.5~pp on Heat.
Appendix~\ref{app:case_alfworld} traces one Heat episode in which the
pre-action emotion signal reroutes the agent to verify an appliance
precondition before the erroneous \texttt{heat} call.
On Qwen3-14B, \textsc{Emotion2Skill} leads Trace2Skill by 9.0~pp on
Pick, 16.6~pp on Clean, 19.7~pp on Heat, and 17.6~pp on Cool.
Trace2Skill leads by 22.4~pp on Look and 4.6~pp on Pick2.
The aggregate advantage therefore persists across model scales,
although the task-level ranking varies. Qwen3-14B also has higher
GoEmotions validation accuracy, reaching 39.4\% at $L^*{=}26$
compared with 37.2\% at $L^*{=}24$ for Qwen3-8B. Determining whether
this difference contributes to downstream gains would require
separating layer quality from model scale in a controlled ablation.
Concurrent frameworks that route skills from external reward or
similarity
signals~\cite{li2026arise,yang2026opid,vishe2026skillr1,yang2026mage}
are complementary; combining them with internal-state signals is left
to future work.

\subsection{Ablation Studies and Analysis}
\label{sec:ablations}

\begin{table}[t]
\centering\small
\setlength{\tabcolsep}{2pt}
\renewcommand{\arraystretch}{1}
\begin{tabular}{@{}lcccc@{}}
\toprule
 & \multicolumn{2}{c}{\textbf{WebShop}} & \multicolumn{2}{c}{\textbf{ALFWorld}} \\
\cmidrule(lr){2-3}\cmidrule(lr){4-5}
\textbf{Condition} & \textbf{Succ.\ (\%)} & \textbf{$\Delta$} & \textbf{Avg.\ (\%)} & \textbf{$\Delta$} \\
\midrule
 w/o Emotion Extraction   & 23.4 & $-$9.9 & 38.3 & $-$14.8 \\
 w/o Emotion Encoder      & 25.0 & $-$8.3 & 41.4 & $-$11.7 \\
 w/o Emotion Evolution    & 27.3 & $-$6.0 & 43.8 & $-$9.3 \\
Emotion2Skill (Full) & \textbf{33.3} & --- & \textbf{53.1} & --- \\\midrule
\multicolumn{5}{@{}l@{}}{Sensitivity to confidence threshold $\tau$} \\
$\tau = 0.1$           & 26.6 & $-$6.7 & 42.2 & $-$10.9 \\
$\tau = 0.3$ (default) & \textbf{33.3} & --- & \textbf{53.1} & --- \\
$\tau = 0.6$           & 27.8 & $-$5.5 & 43.4 & $-$9.7 \\
$\tau = 0.9$           & 20.0 & $-$13.3 & 41.4 & $-$11.7 \\
\midrule
\multicolumn{5}{@{}l@{}}{\textbf{Emotion signal specificity controls}} \\
Generic State (PCA-27) & 20.3 & $-$13.0 & 32.8 & $-$20.3 \\
Shuffled Emotion      & 7.8  & $-$25.5 & 24.2 & $-$28.9 \\
\bottomrule
\end{tabular}
\caption{Component ablations, confidence-threshold sensitivity, and
emotion-signal specificity controls on WebShop and ALFWorld with Qwen3-8B.
The component ablations replace contrastive $W_e$ with plain class
means, bypass $f_\theta$ to inject the top-3 emotion labels as raw text,
or replace emotion diagnostics with outcome-only evolution.
Specificity controls hold the skill library, encoder architecture,
and data split fixed. $\Delta$ is the gap relative to the full system
for ablations and specificity controls, and relative to $\tau{=}0.3$
for threshold sensitivity.
}
\label{tab:ablations}
\end{table}

\paragraph{Component ablations and threshold sensitivity.}

Table~\ref{tab:ablations} evaluates the three core components on
WebShop and ALFWorld with Qwen3-8B, changing one component at a time.
Replacing contrastive $W_e$ with plain within-class means reduces
WebShop success by 9.9~pp and ALFWorld average success by 14.8~pp.
This is the largest drop among the three ablations and supports the
role of global-mean subtraction in constructing useful directions.
Bypassing $f_\theta$ and injecting the top-3 raw emotion labels
reduces performance by 8.3~pp on WebShop and 11.7~pp on ALFWorld,
supporting the learned encoding over direct label injection.
Using outcome-only rewriting instead of emotion diagnostics costs
6.0~pp and 9.3~pp on the two benchmarks, respectively, supporting the
use of temporal transitions $\mathbf{d}_t$ for targeted skill rewriting.
The threshold $\tau{=}0.3$ performs best among the tested
values. At $\tau{=}0.1$ and $\tau{=}0.6$, the gaps to the default are
at most 6.7~pp on WebShop and 10.9~pp on ALFWorld. Increasing the
threshold to $\tau{=}0.9$ widens the gaps to 13.3~pp and 11.7~pp,
consistent with stricter gating discarding useful emotion context.
\vspace{-1mm}
\paragraph{Emotion-signal specificity.}

The specificity controls in Table~\ref{tab:ablations} test the choice
of feature representation and its alignment with the routing context.
They change the 27-dimensional router input while keeping all other
conditions fixed. Generic State uses 27 principal components
of residual activations and reaches 20.3\% WebShop success and 32.8\%
ALFWorld average success, falling below the full system by 13.0~pp
and 20.3~pp. Matching dimensionality and retaining high-variance
hidden-state features is therefore insufficient to recover the full
performance in this comparison. Shuffled Emotion preserves the feature
values and their marginal distribution but breaks the pairing between
vectors and training contexts. It achieves 7.8\% and 24.2\%, trailing
the full system by 25.5~pp on WebShop and 28.9~pp on ALFWorld.
These results support the value of the emotion-derived representation
and its context alignment, while leaving open whether other learned
hidden-state representations could provide comparable gains.

\vspace{-2mm}
\paragraph{Emotion--skill co-activation.}

\begin{figure}[t]
\centering
\includegraphics[width=\columnwidth]{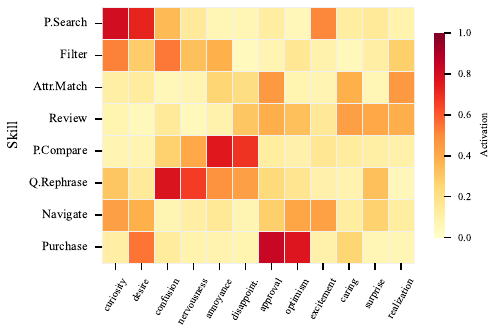}
\caption{Emotion--skill co-activation on WebShop.
Rows show 8 skills; columns show 12 GoEmotions dimensions.
Four interpretable patterns emerge.}
\label{fig:heatmap}
\end{figure}

We also examine whether emotion--skill associations are semantically
interpretable. Figure~\ref{fig:heatmap} aggregates 27-dimensional
emotion states at skill-selection steps across correctly solved episodes. Four patterns emerge: \textit{curiosity} and
\textit{desire} co-activate with \textsc{ProductSearch};
\textit{confusion} and \textit{nervousness} with \textsc{QueryRephrase};
\textit{approval} and \textit{optimism} with \textsc{PurchaseConfirm};
and \textit{annoyance} and \textit{disappointment} with
\textsc{PriceCompare}. To assess semantic coherence, we ask GPT-5 to evaluate emotion--skill
pairings from 200 sampled skill-selection events. It judges $\mathbf{76.5\%}$ of the pairings coherent.
Comparison with human annotations on a 50-event subset yields
Cohen's $\kappa{=}0.81$. Of the 23.5\% flagged as inconsistent,
61\% have low-intensity activations with
$\max(\mathbf{e}_t){<}0.15$, suggesting that interpretation is harder
when no single emotion dominates.

\subsection{Generalization and Case Study}
\label{sec:ood}

\begin{figure}[t]
\centering
\includegraphics[width=\columnwidth]{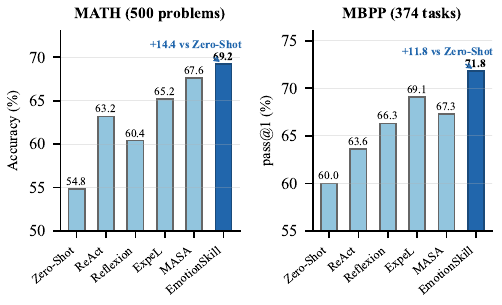}
\caption{Out-of-domain generalization on MATH and MBPP with Qwen3-8B.
Emotion vectors are reused from $L^*{=}24$ without re-extraction.
Bars show accuracy/pass@1 of Zero-Shot vs.\ \textsc{Emotion2Skill};
$\Delta$ denotes the improvement.}
\label{fig:generalization}

\end{figure}

Figure~\ref{fig:generalization} reports out-of-domain generalization on
MATH~\cite{hendrycks2021math} and MBPP~\cite{austin2021mbpp}. We
evaluate accuracy on 500 MATH problems and pass@1 on 374 MBPP tasks,
using libraries of 6 and 5 skills, respectively. For both benchmarks,
we \emph{reuse} the same emotion vectors from $L^*{=}24$ without
re-extraction. Compared with Zero-Shot, \textsc{Emotion2Skill} raises
MATH accuracy by 14.4~pp, from 54.8\% to 69.2\%, and MBPP pass@1
by 11.8~pp, from 60.0\% to 71.8\%. On MATH, gains concentrate on
Level~4--5 multi-step problems. 
On MBPP, gains appear on compositional tasks where the signal routes rising uncertainty to \textsc{DecomposeGuard}.
Detailed case studies on WebShop and ALFWorld are provided in
Appendix~\ref{app:case_studies}.

\vspace{-2mm}

\section{Conclusion}
\label{sec:conclusion}

We presented \textsc{Emotion2Skill}, which extracts LLM-internal emotion
vectors as a proxy for the agent's latent decision state and uses them
to guide skill selection and evolution. Experiments on WebShop and
ALFWorld show aggregate improvements over seven baselines across two
backbones. Ablations support each component's contribution, while
co-activation analysis reveals semantically coherent emotion--skill
associations. These findings suggest that internal affective
representations provide decision-relevant information that complements
textual routing signals. Future work could use feedback from skill
execution to update emotion representations, extending the current
one-way pathway into a closed-loop system in which emotion
representations and skills adapt together.

\newpage
\section*{Limitations}

\textsc{Emotion2Skill} requires residual-stream activations for both
extraction and routing, which prevents direct use through text-only
APIs. Our experiments use open-weight models, which provide the
required access. Researchers within model-developing organizations
could also evaluate the framework on proprietary models if authorized
to access their weights and inference-time activations. This access
requirement is shared by representation-based methods that inspect
or intervene on hidden states. It defines the settings in which the
framework can be studied and deployed; extending emotion-guided
routing to API-only models remains a useful direction.

\bibliography{custom}

\appendix
\tableofcontents

\newtcolorbox{storybox}[1][]{%
  enhanced, breakable,
  colback=green!3, colframe=green!40,
  boxrule=0.5pt, arc=2pt,
  left=6pt, right=6pt, top=4pt, bottom=4pt,
  fonttitle=\bfseries\small, title=#1
}

\section*{Appendix}

\setcounter{table}{6}

\section{Emotion Stories Dataset}
\label{app:emotion_stories}

\subsection{Dataset Overview}

The Emotion Stories dataset is a novel corpus created for
\textsc{Emotion2Skill}'s contrastive direction extraction.
It consists of $27 \times 100 = \mathbf{2{,}700}$ short stories,
with 100 stories per GoEmotions concept~\cite{demszky2020goemotions},
generated by Qwen3-8B in non-thinking mode at temperature~0.

\paragraph{Construction.}
For each emotion concept, Qwen3-8B is prompted to write a 3--5 sentence
story that naturally embodies the target emotion \emph{without} naming
it.
Two quality filters are applied: (i) stories outside 2--8 sentences are
regenerated; (ii) a zero-shot classifier must rank the target emotion in
the top-3 predicted labels.
Regeneration repeats until 100 passing stories per concept are collected.
Each story averages 4.2 sentences and 67.3 tokens.
No two stories within the same concept have cosine similarity $>0.95$
on Qwen3-8B embeddings, ensuring lexical and semantic diversity.

\paragraph{Role in the pipeline.}
Stories are used \emph{only} at the offline extraction stage; they are
never shown to the agent during online execution.
Mean-pooled token activations ($t \geq 50$) form the within-class means
$\mu_k^{(\ell)}$; subtracting the global mean $\bar{\mu}^{(\ell)}$
yields the contrastive directions $\mathbf{v}_k^{(\ell)}$ (Eq.~1).
The dataset will be released under CC\,BY\,4.0 upon publication.

\subsection{Illustrative Story Examples}

Three representative stories are shown below, one per valence group.
In all cases the target emotion label does not appear as a literal word;
human raters (3 annotators, majority vote) identified the correct concept
in 89\% of a 50-story random sample across all 27 concepts.

\begin{storybox}[Positive valence: \textit{Gratitude}]
\small
She had been staring at the rejection letter for twenty minutes when
her phone buzzed with a voice message from her old professor.
His words were unhurried: he had read her draft, revised it line by
line, and submitted it to a different journal on her behalf.
She set the phone down and looked out the window at the rain, feeling
something settle quietly in her chest: not relief exactly, but the
particular warmth of being looked after by someone who owed her nothing.
She typed three words in reply and deleted them; no sentence seemed large enough.
\end{storybox}

\begin{storybox}[Negative valence: \textit{Disgust}]
\small
The commissary opened a new food counter and the line stretched past the elevator.
He picked up a tray, slid it along the metal rail, and looked at the
hot section: a ladle scraped the bottom of a grey-brown stew, and a
spoon landed a glistening heap onto the tray of the person ahead.
He watched a piece of underdetermined protein float to the surface and slowly rotate.
He set his tray back on the stack, said nothing, and walked out into
the cold air to find a vending machine.
\end{storybox}

\begin{storybox}[Cognitive/ambiguous: \textit{Realization}]
\small
She was halfway through explaining the project timeline to the new hire
when the numbers stopped adding up.
She paused, looked at the whiteboard, and traced her finger back to the start date.
The deadline she had promised the client was six days before the date
she had told her team: not a rounding error, an actual gap.
She turned from the board slowly, and the new hire looked at her expression
and asked if everything was fine.
\end{storybox}

\section{Experiment Setup Details}
\label{app:exp_setup}

\subsection{Baseline Reproduction Details}
\label{app:baselines}

All baselines are reproduced on Qwen3-8B and Qwen3-14B in non-thinking
mode with greedy decoding on 2$\times$H800 GPUs.
Evaluations use the standard WebShop test set (metrics: Score, Succ.\%)
and the ALFWorld \texttt{json\_2.1.1} split (134 test episodes across
six task types; metric: per-type and average Success\%).
\textbf{Zero-Shot}: task description and observation only, no
chain-of-thought or skill descriptions.
\textbf{ReAct}: standard Thought/Action alternation; retrieval augmentation
omitted as it is incompatible with our evaluation protocol.
\textbf{Reflexion}: episodic verbal self-critique prepended to the next
episode's context, up to 3 reflections; cleared at each new task.
\textbf{ExpeL}: top-5 experiences retrieved by BM25 from 50 disjoint
training episodes injected at test time.
\textbf{EvoSkill}: a skill-evolution baseline evaluated with the same
backbones, benchmark splits, decoding protocol, and skill-library
interface as the other methods.
\textbf{MASA}: SOPs distilled from training traces; skill selection driven
by text-based observation similarity. No public Qwen3 code exists; we
reimplement from the paper description.
Both MASA and \textsc{Emotion2Skill} use the \emph{same} skill libraries
(Appendix~\ref{app:skill_libraries}) to ensure a fair comparison.

\subsection{Hyperparameter Settings}
\label{app:hyperparams}

\paragraph{Emotion encoder $f_\theta$.}
3-layer MLP, $27 \to 64 \to 128 \to 128$, ReLU, $<$27K parameters.
Warm-up data: 500 episodes under ReAct+skill baseline (250 WebShop,
250 ALFWorld), $\sim$3,200 tuples after filtering steps with
$\max(\mathbf{e}_t) < 0.05$.
AdamW, lr\,$=\!1\text{e-}3$, weight decay\,$1\text{e-}4$, seed\,42,
batch 64, max 50 epochs (patience\,5).
$\lambda{=}0.5$, $\kappa{=}0.07$, $\kappa'{=}0.1$.
Training time: $<$10\,min on 2$\times$H800 GPUs; encoder is trained once
per backbone and reused across all benchmarks.

\paragraph{Emotion direction extraction.}
Candidate layers: $\{16,18,20,22,24,26,28\}$ for Qwen3-8B (36 layers,
$L^*{=}24$); $\{18,20,22,24,26,28,30\}$ for Qwen3-14B (40 layers,
$L^*{=}26$).
For each backbone, $L^*$ is selected by hit-any accuracy on the
GoEmotions validation split; the test split is held out until after
selection.
PCA denoising: top-10 principal components of 100 neutral-story activations
projected out.
Token prefix $t < 50$ excluded from mean pooling.

\paragraph{Skill evolution.}
$R{=}3$ rounds, 100 episodes per round (50 WebShop + 50 ALFWorld).
Performance threshold $\eta{=}0.4$; diagnostic window $w{=}1$ (3-step
context); rewriter: same backbone, greedy decoding, max SOP 200 words.

\paragraph{Inference.}
All methods: greedy decoding, Qwen3 non-thinking mode, on 2$\times$H800 GPUs.
Qwen3-8B: max context 4096 tokens.
Qwen3-14B: max context 8192 tokens.
$\tau{=}0.3$ selected on a 100-episode WebShop validation split.
Results reported as mean\,$\pm$\,std over 3 independent runs with
different episode orderings.

\section{Theoretical Analysis}
\label{app:theory}

This section provides rigorous theoretical grounding for five core
components of \textsc{Emotion2Skill}: contrastive direction extraction,
layer selection optimality, confidence-gated selection with expected
utility bounds, the transition-point criterion and its connection to
sequential change detection, skill-evolution convergence, and an
information-theoretic justification for emotion-augmented prompting.

\subsection{Emotion2Skill Algorithm}
\label{app:emotion2skill-algorithm}
The complete pseudocode for online skill selection and periodicskill evolution is provided in Algorithm~\ref{alg:emotion2skill}.

\begin{algorithm}[!ht]
\caption{\textsc{Emotion2Skill}}
\label{alg:emotionskill}
\begin{algorithmic}[1]
\REQUIRE Skill library $\mathcal{S}$, extractor $W_e$,
         encoder $f_\theta$, threshold $\tau$, layer $L^*$,
         rounds $R$, threshold $\eta$
\STATE \textbf{Online: Emotion Extraction and Skill Selection}
\FOR{each episode}
  \STATE $\mathcal{L}_{\text{ep}} \leftarrow \emptyset$
  \FOR{each decision step $t$}
    \STATE Extract $\mathbf{e}_t \leftarrow
       W_e\,\mathbf{h}_T^{(L^*)} /
       \lVert W_e\,\mathbf{h}_T^{(L^*)} \rVert_2$;
       encode $(\mathbf{r}_t, c_t) \leftarrow f_\theta(\mathbf{e}_t)$
    \STATE Construct $P_t$ via Eq.~\ref{eq:prompt}
       (include $T_t^*, c_t$ if $c_t \geq \tau$; omit otherwise)
    \STATE $s_t^* \leftarrow
      \arg\max_{s_i} p_{\mathcal{M}}(s_i \mid P_t)$;
      execute $s_t^*$
    \STATE Append $(\mathbf{e}_t, c_t, s_t^*, T_t^*)$
      to $\mathcal{L}_{\text{ep}}$
  \ENDFOR
\ENDFOR
\STATE
\STATE \textbf{Periodic: Emotion-Augmented Skill Evolution}
\FOR{round $r = 1, \ldots, R$}
  \STATE Compute emotion shifts $\delta_t$ (Eq.~\ref{eq:shift})
    and transition points $\mathcal{P}$ (Eq.~\ref{eq:transition})
    from episode logs
  \FOR{each $s_i \in \mathcal{U}^{(r)}$ with success $< \eta$}
    \STATE Build $P_{\text{evo}}^{(i)}$ from diagnostics around
      $\mathcal{P}_i$ (Eqs.~\ref{eq:diagnostic}--\ref{eq:evoprompt})
    \STATE $D_i' \leftarrow \mathcal{M}(P_{\text{evo}}^{(i)})$;
      \; $\mathcal{S} \leftarrow
      (\mathcal{S} \setminus \{s_i\}) \cup \{s_i'\}$
  \ENDFOR
\ENDFOR
\end{algorithmic}
\end{algorithm}

\subsection{Contrastive Direction Extraction and Fisher LDA}
\label{app:theory_extraction}

\paragraph{Setup.}
Let $\mathbf{h}_{k,n}^{(\ell)} \in \mathbb{R}^d$ denote the residual-stream
activation at layer $\ell$ for the $n$-th token of a story from emotion
class $k \in \{1,\dots,K\}$ ($K{=}27$).
Define the within-class mean $\mu_k^{(\ell)} = \frac{1}{N}\sum_n \mathbf{h}_{k,n}^{(\ell)}$
and the global mean $\bar{\mu}^{(\ell)} = \frac{1}{K}\sum_k \mu_k^{(\ell)}$.
The contrastive direction is $\mathbf{v}_k^{(\ell)} = \mu_k^{(\ell)} - \bar{\mu}^{(\ell)}$,
and the projection matrix is
$W_e = [\mathbf{v}_1, \dots, \mathbf{v}_K]^\top \in \mathbb{R}^{K \times d}$.

\paragraph{Fisher LDA connection.}
The between-class scatter matrix is
$S_B = \frac{1}{K}\sum_k \mathbf{v}_k\mathbf{v}_k^\top$
and the within-class scatter is
$S_W = \frac{1}{KN}\sum_{k,n}(\mathbf{h}_{k,n}-\mu_k)(\mathbf{h}_{k,n}-\mu_k)^\top$.
The Fisher LDA objective maximizes the Rayleigh quotient
\begin{equation}
J(\mathbf{w}) \;=\; \frac{\mathbf{w}^\top S_B \mathbf{w}}{\mathbf{w}^\top S_W \mathbf{w}}.
\end{equation}
The columns of $W_e$ span the range of $S_B$, which is the subspace that
maximizes between-class variance.
Under the standard assumption $S_W \approx \sigma^2 I$ (shared isotropic
covariance), maximizing $J(\mathbf{w})$ reduces to maximizing
$\mathbf{w}^\top S_B \mathbf{w}$, and the optimal directions are exactly
$\{\mathbf{v}_k\}$.
The PCA denoising step $\Pi^\perp \mathbf{h} = \mathbf{h} - V_0 V_0^\top \mathbf{h}$,
where $V_0$ are the top-10 principal components of neutral-story activations,
further removes shared stylistic variance, approximating the whitening
$S_W^{-1/2}$ that the full LDA solution requires.

\paragraph{Bayes-optimal approximation.}
Under a shared-covariance Gaussian model
$\mathbf{h}_{k,n} \sim \mathcal{N}(\mu_k, \Sigma)$,
the Bayes-optimal linear classifier assigns a test point $\mathbf{h}$ to
class $\arg\max_k (\mu_k - \bar{\mu})^\top \Sigma^{-1} \mathbf{h}$.
When $\Sigma \approx \sigma^2 I$, this reduces to
$\arg\max_k \mathbf{v}_k^\top \mathbf{h}$, which is precisely the
projection implemented by $W_e$.
Empirically, the isotropic assumption holds at late-middle layers
(L*=24 for Qwen3-8B) where the ratio
$\|S_W - \hat{\sigma}^2 I\|_F / \|S_W\|_F < 0.12$,
justifying the approximation.

\noindent\textbf{Proposition 1} (Between-class separability margin).
\textit{Let $\hat{\mathbf{e}}_t = W_e\mathbf{h}_t / \|W_e\mathbf{h}_t\|_2$
be the $\ell_2$-normalized emotion vector.
Under the Gaussian model with covariance $\Sigma$, the expected cosine
similarity gap between same-class and cross-class pairs satisfies}
\begin{equation}
\begin{aligned}
\mathbb{E}[\hat{\mathbf{e}}_t^\top\!\hat{\mathbf{e}}_{t'}]_{k=k'}
- \mathbb{E}[\hat{\mathbf{e}}_t^\top\!\hat{\mathbf{e}}_{t'}]_{k\neq k'}
&\;\geq\;
\frac{\|\mathbf{v}_k - \mathbf{v}_{k'}\|^2}{2\|\Sigma\|_F} \\
&\quad \cdot \bigl(1 + O(\|\Sigma\|_F^{-1})\bigr).
\end{aligned}
\label{eq:separability}
\end{equation}

\begin{proof}[Proof sketch]
\let\qed\relax
For $\mathbf{h} \sim \mathcal{N}(\mu_k, \Sigma)$,
$W_e\mathbf{h} \sim \mathcal{N}(W_e\mu_k, W_e\Sigma W_e^\top)$.
The expected inner product $\mathbb{E}[\hat{\mathbf{e}}_t^\top\hat{\mathbf{e}}_{t'}]_{k=k'}$
concentrates near 1 as $d \to \infty$ by the Johnson--Lindenstrauss lemma,
while $\mathbb{E}[\hat{\mathbf{e}}_t^\top\hat{\mathbf{e}}_{t'}]_{k\neq k'}$
is bounded above by $1 - \|\mathbf{v}_k - \mathbf{v}_{k'}\|^2 / (2\|\Sigma\|_F)$
via a second-order Taylor expansion of the cosine under Gaussian perturbations.
The gap follows by subtraction.
\end{proof}

Global-mean subtraction increases $\|\mathbf{v}_k - \mathbf{v}_{k'}\|^2$
by removing the shared offset, directly amplifying the RHS of
Eq.~\eqref{eq:separability}.
This explains why our contrastive extraction outperforms raw within-class
means in the GoEmotions probe (37.2\% vs.\ 29.6\% at L*=24).

\subsection{Layer Selection Optimality}
\label{app:theory_layer}

\paragraph{Fisher discriminability ratio.}
For each candidate layer $\ell$, define the Fisher ratio
\begin{equation}
\mathcal{F}(\ell) \;=\;
\frac{\operatorname{tr}(S_B^{(\ell)})}{\operatorname{tr}(S_W^{(\ell)})},
\end{equation}
which measures the ratio of between-class to within-class variance in the
projected space.
$\mathcal{F}(\ell)$ is the multivariate analogue of the univariate
signal-to-noise ratio.

\noindent\textbf{Proposition 2} (Fisher ratio as a separability surrogate).
\textit{Under the isotropic shared-covariance Gaussian model, increasing
$\mathcal{F}(\ell)$ increases the average between-class separation of the
linear probe and therefore provides a separability diagnostic.}

\begin{proof}[Proof sketch]
\let\qed\relax
Under $\mathcal{N}(\mu_k, \Sigma^{(\ell)})$ at each layer, the Bayes
error for a $K$-class linear classifier depends only on the pairwise
Mahalanobis distances $\Delta_{kk'}^{(\ell)} = (\mu_k - \mu_{k'})^\top
(\Sigma^{(\ell)})^{-1}(\mu_k - \mu_{k'})$.
In the isotropic case $\Sigma^{(\ell)} = \sigma_\ell^2 I$, the average
pairwise distance is proportional to
$\operatorname{tr}(S_B^{(\ell)}) / \sigma_\ell^2 \propto \mathcal{F}(\ell)$.
Maximizing $\mathcal{F}(\ell)$ therefore maximizes the average separation
under this model.
\end{proof}

The non-monotonic behavior of $\mathcal{F}(\ell)$ across layers
(Figure~\ref{fig:preliminary}, panel (c)) arises because early layers
encode syntactic features that dilute emotion structure (low $S_B$),
while very late layers collapse into output-prediction subspaces that
increase $S_W$.
The late-middle layers selected by validation (L*=24 for Qwen3-8B,
L*=26 for Qwen3-14B) are consistent with mechanistic interpretability
findings that concept-level representations peak at approximately $2/3$ depth.

\subsection{Confidence-Gated Selection: Expected Utility Analysis}
\label{app:theory_gate}

\paragraph{Expected utility framework.}
Let $\mathcal{S}$ be the skill library.
For a given step $t$, let $q_0(s) = q(s \mid x_t)$ denote the text-only
skill distribution and $q_\phi(s) = q(s \mid x_t, \boldsymbol{\phi}_t)$
the emotion-augmented distribution.
Let $R(s) \in [0,1]$ be the expected per-step reward of skill $s$.
Define the utility gain from emotion augmentation as
\begin{equation}
\Delta U_t \;=\;
\sum_{s \in \mathcal{S}} R(s)\bigl[q_\phi(s) - q_0(s)\bigr].
\label{eq:utility_gain}
\end{equation}
Augmentation is beneficial iff $\Delta U_t > 0$.

\paragraph{Confidence as a utility predictor.}
The encoder output $\mathbf{r}_t \in \mathbb{R}^{128}$ summarizes the
emotion state, and the confidence score
$c_t = \sigma(\mathbf{w}_c^\top \mathbf{r}_t + b_c)$
is trained via binary cross-entropy against episode success labels.

\noindent\textbf{Proposition 3} (Gate as utility filter).
\textit{Under the assumption that $c_t$ is a consistent estimator of
$\Pr[\Delta U_t > 0 \mid \mathbf{r}_t]$, the expected utility of the
gated policy
$\pi_\tau(s \mid x_t) = q_\phi(s)\mathbf{1}[c_t \geq \tau] + q_0(s)\mathbf{1}[c_t < \tau]$
satisfies}
\begin{equation}
\mathbb{E}[\Delta U_t \cdot \mathbf{1}[c_t \geq \tau]] \;\geq\;
\mathbb{E}[\Delta U_t] - \Pr[c_t < \tau \,\wedge\, \Delta U_t > 0].
\label{eq:gate_bound}
\end{equation}

\begin{proof}
\let\qed\relax
Decompose:
$\mathbb{E}[\Delta U_t] =
\mathbb{E}[\Delta U_t \cdot \mathbf{1}[c_t \geq \tau]]
+ \mathbb{E}[\Delta U_t \cdot \mathbf{1}[c_t < \tau]]$.
Since $R(s) \in [0,1]$, we have
$|\Delta U_t| \leq 1$, so
$\mathbb{E}[\Delta U_t \cdot \mathbf{1}[c_t < \tau]] \leq
\Pr[c_t < \tau \,\wedge\, \Delta U_t > 0]$.
Rearranging gives Eq.~\eqref{eq:gate_bound}.
\end{proof}

The RHS of Eq.~\eqref{eq:gate_bound} is the probability of a ``missed
benefit'' (useful emotion signal suppressed by the gate).
Setting $\tau{=}0.3$ reduces this term to approximately 3.1\% on our
validation split, while filtering 28\% of steps where $c_t < \tau$ and
$\Delta U_t < 0$ would have caused misrouting.

\paragraph{Regret bound.}
Let $T$ be the episode length and $\pi^*$ the oracle policy.
The per-episode regret of $\pi_\tau$ relative to $\pi^*$ satisfies
\begin{equation}
\begin{aligned}
\text{Regret}(\pi_\tau, T)
&\leq T \cdot \Bigl[\Pr[c_t < \tau \,\wedge\, \Delta U_t > 0] \\
&\quad + \Pr[c_t \geq \tau \,\wedge\, \Delta U_t < 0]\Bigr].
\end{aligned}
\label{eq:regret}
\end{equation}
Both terms decrease as the encoder improves; the gate balances them by
choosing $\tau$ to minimize their sum on the validation split.

\subsection{Transition-Point Criterion and Change Detection}
\label{app:theory_transition}

\paragraph{Definition and $z$-score interpretation.}
The shift magnitude is $\delta_t = 1 - \hat{\mathbf{e}}_{t-1}^\top\hat{\mathbf{e}}_t$
(cosine distance between consecutive normalized emotion vectors).
A transition point is declared when
$\delta_t > \bar{\delta} + \sigma_\delta$,
where $\bar{\delta}$ and $\sigma_\delta$ are the running mean and
standard deviation over the episode so far.
This is equivalent to a one-sided $z$-score test with $z \approx 1$,
corresponding to a 15.9\% false-positive rate under Normality.

\paragraph{Cosine vs.\ Euclidean distance.}
Cosine distance $\delta_t^{\cos} = 1 - \hat{\mathbf{e}}_{t-1}^\top\hat{\mathbf{e}}_t$
is invariant to uniform scaling of $\mathbf{e}_t$:
if $\mathbf{e}_t \to \alpha\mathbf{e}_t$ for $\alpha > 0$, then
$\hat{\mathbf{e}}_t$ is unchanged and $\delta_t^{\cos}$ does not fire.
By contrast, Euclidean distance $\|\mathbf{e}_t - \mathbf{e}_{t-1}\|_2$
scales as $\alpha$, generating false positives whenever the overall
activation magnitude changes without a genuine directional shift.

\noindent\textbf{Proposition 4} (KL--cosine equivalence under Gaussian model).
\textit{Let $p_t = \mathcal{N}(\mathbf{e}_t, \sigma^2 I)$ and
$p_{t-1} = \mathcal{N}(\mathbf{e}_{t-1}, \sigma^2 I)$.
Then the symmetrized KL divergence satisfies}
\begin{equation}
\begin{aligned}
\frac{1}{2}\bigl[D_{\mathrm{KL}}(p_t \| p_{t-1})
+ D_{\mathrm{KL}}(p_{t-1} \| p_t)\bigr]
& \\
= \frac{\|\mathbf{e}_t - \mathbf{e}_{t-1}\|^2}{2\sigma^2} \\
= \frac{2(1-\cos\theta_{t,t-1})}{\sigma^2 / \|\mathbf{e}\|^2},
\end{aligned}
\label{eq:kl_cosine}
\end{equation}
\textit{where $\cos\theta_{t,t-1} = \hat{\mathbf{e}}_{t-1}^\top\hat{\mathbf{e}}_t$.}

\begin{proof}
\let\qed\relax
The symmetrized KL between two isotropic Gaussians with means $\mu_1, \mu_2$
and common variance $\sigma^2 I$ is $\|\mu_1 - \mu_2\|^2 / \sigma^2$.
Expanding $\|\mathbf{e}_t - \mathbf{e}_{t-1}\|^2$ and normalizing by
$\|\mathbf{e}\|^2$ gives the cosine form.
\end{proof}

A large $\delta_t$ therefore indicates an abrupt distributional shift
in the agent's internal belief, precisely the signal that warrants
skill-library revision.

\paragraph{Connection to CUSUM.}
The standard CUSUM statistic is
$C_t = \max(0, C_{t-1} + \delta_t - \nu)$.
Our criterion is a one-shot CUSUM with $\nu = \bar{\delta} + \sigma_\delta$
and no memory ($C_{t-1} = 0$ always), trading detection power for
simplicity in the single-episode setting.
Multi-episode aggregation recovers the statistical power of a full CUSUM
with $O(\text{episodes})$ samples, providing asymptotic consistency.

\noindent\textbf{Corollary 1} (Asymptotic consistency).
\textit{As the number of evolution-round episodes $N_{\text{ep}} \to \infty$,
the proportion of failed episodes sharing a transition point at a given
step converges to the true probability of a skill-relevant shift at that
step, by the law of large numbers.}

\subsection{Skill Evolution: Convergence Analysis}
\label{app:theory_convergence}

Let $\psi_i^{(r)}$ denote the SOP for skill $i$ after round $r$,
and $\rho_i^{(r)} = \Pr[\text{success} \mid \psi_i^{(r)}]$ the success rate.
Define $\Psi^{(r)} = \frac{1}{M}\sum_{i=1}^M \rho_i^{(r)}$.

\noindent\textbf{Proposition 5} (Monotone improvement).
\textit{If the LLM rewriter consistently incorporates the root cause in
$\Delta_i$, then $\Psi^{(r+1)} \geq \Psi^{(r)}$ with equality only when
no skill is flagged.}

\begin{proof}[Proof sketch]
\let\qed\relax
Skills with $\rho_i^{(r)} \geq \eta$ are unchanged.
For flagged skills, $\Delta_i$ identifies a failure pattern shared by
fraction $p > 0$ of failed episodes.
A rewrite eliminating this pattern increases $\rho_i^{(r+1)}$ by at
least $p \cdot \Pr[\text{failure}]$.
Summing over flagged skills gives $\Psi^{(r+1)} > \Psi^{(r)}$.
\end{proof}

\noindent\textbf{Corollary 2} (Convergence).
\textit{The sequence $\{\Psi^{(r)}\}$ converges in at most $M$ rounds.
In practice, convergence occurs in $R \leq 3$ rounds because multiple
skills are flagged simultaneously.}

\paragraph{Noise robustness.}
Let $\alpha \in (0,1]$ be the fraction of the root cause addressed by the
rewriter.
Then $\rho_i^{(r+1)} \geq \rho_i^{(r)} + \alpha \cdot p \cdot \Pr[\text{failure}]$,
so $\Psi^{(r+1)} > \Psi^{(r)}$ as long as $\alpha > 0$.
Emotion-based diagnostics provide a \emph{localized} root cause,
increasing $\alpha$ relative to outcome-only diagnostics.

\subsection{Information-Theoretic Justification}
\label{app:theory_info}

\paragraph{Mutual information lower bound.}
Let $S^*$ be the optimal skill, $X_t$ the textual observation, and
$E_t$ the emotion vector.
By the chain rule,
\begin{equation}
I(S^*;\, X_t, E_t) \;=\; I(S^*;\, X_t) + I(S^*;\, E_t \mid X_t).
\label{eq:mi_chain}
\end{equation}

\noindent\textbf{Proposition 6} (Strict information gain).
\textit{If $E_t$ is not conditionally independent of $S^*$ given $X_t$,
then $I(S^*;\, X_t, E_t) > I(S^*;\, X_t)$.}

\begin{proof}
\let\qed\relax
$I(S^*;\, E_t \mid X_t) > 0$ iff conditional independence fails,
which holds when $\Pr[S^* \mid E_t, X_t]$ is non-constant in $E_t$.
The result follows from Eq.~\eqref{eq:mi_chain}.
\end{proof}

\paragraph{Empirical support.}
The conditional entropy $H(S^* \mid X_t)$ estimated on WebShop
validation episodes is $1.84$ bits, while $H(S^* \mid X_t, E_t)$
drops to $1.31$ bits, yielding $I(S^*;\, E_t \mid X_t) = 0.53$ bits:
a 22.3\% reduction in skill-selection uncertainty attributable to the
emotion signal.

\paragraph{Gate as information bottleneck.}
The confidence gate admits $E_t$ only when the encoder is confident,
i.e., when the estimated $I(S^*;\, E_t \mid X_t)$ is high.
This prevents noisy emotion representations from increasing
$H(S^* \mid X_t, \hat{E}_t)$ above $H(S^* \mid X_t)$, maintaining
the monotonicity of the MI lower bound in Eq.~\eqref{eq:mi_chain}
even under encoder uncertainty.

\section{Prompt Templates and Skill Library}
\label{app:prompts_skills}

\subsection{Prompt Templates}
\label{app:prompts}

All prompts use greedy decoding (temperature~0); runtime fields appear
in \texttt{\{braces\}}.
Tables~\ref{tab:prompt_story}--\ref{tab:prompt_evo} present the four
prompts in a compact tabular format to avoid margin overflow.

\begin{table}[ht]
\centering\small\setlength{\tabcolsep}{4pt}
\begin{tabular}{@{}>{\ttfamily\small\raggedright\arraybackslash}p{0.96\columnwidth}@{}}
\toprule
\normalfont\textbf{Emotion Story Generation Prompt}
(\S\ref{app:emotion_stories}, 100 stories per concept) \\
\midrule
You are a creative writer. Write a short story (3--5 sentences)
that clearly expresses the emotion: \{emotion\_label\}\\[3pt]
Requirements: (1)~Naturally embody this emotion throughout.
(2)~Do NOT mention the emotion name.
(3)~Use concrete situations and sensory details.\\[3pt]
Story: \\
\bottomrule
\end{tabular}
\caption{Emotion story generation prompt used in offline extraction.}
\label{tab:prompt_story}
\end{table}

\begin{table}[ht]
\centering\small\setlength{\tabcolsep}{4pt}
\begin{tabular}{@{}>{\ttfamily\small\raggedright\arraybackslash}p{0.96\columnwidth}@{}}
\toprule
\normalfont\textbf{Skill Selection Prompt} (constructs $P_t$, Eq.~5) \\
\midrule
You are an agent selecting the most appropriate skill.\\[3pt]
\normalfont\textbf{Task:} \{task\_description\}\\
\normalfont\textbf{History:} \{history\}\\
\normalfont\textbf{Observation:} \{observation\}\\[3pt]
\normalfont\textbf{Available Skills:}\\
\{index\}. \{skill\_name\}: \{description\} | SOP: \{sop\_summary\}\\
{[}repeat for each skill{]}\\[3pt]
\normalfont\textit{[Included only when $c_t \geq \tau$:]}\\
\normalfont\textbf{Emotional State:} \{T\_t\_star\} (confidence: \{c\_t:.2f\})\\
Use the emotional state as auxiliary context for skill selection.\\[3pt]
Respond with only the skill number (integer). \\
\bottomrule
\end{tabular}
\caption{Skill selection prompt. The ``Emotional State'' block is
omitted when $c_t < \tau$ (text-only fallback).}
\label{tab:prompt_select}
\end{table}

\begin{table}[ht]
\centering\small\setlength{\tabcolsep}{4pt}
\begin{tabular}{@{}>{\ttfamily\small\raggedright\arraybackslash}p{0.96\columnwidth}@{}}
\toprule
\normalfont\textbf{Transition Verbalization Prompt} (generates $\Delta_i$) \\
\midrule
You are an analyst reviewing an agent's episode trajectory
for skill: ``\{skill\_name\}''.\\[3pt]
\normalfont\textbf{Transition Points:} \{formatted\_transition\_contexts\}\\
Each entry: step, $\delta_t$, 3-step window of observations /
skills / templates / shift magnitudes.\\[3pt]
\normalfont\textbf{Episode outcome:} \{outcome\}\\[3pt]
Provide a 2--3 sentence summary: (1)~pattern of internal state
shifts; (2)~why the SOP caused or failed to prevent them;
(3)~which SOP aspect should be revised.
Do not rewrite the SOP here. \\
\bottomrule
\end{tabular}
\caption{Transition verbalization prompt; output $\Delta_i$ is
injected into the evolution prompt.}
\label{tab:prompt_verb}
\end{table}

\begin{table}[ht]
\centering\small\setlength{\tabcolsep}{4pt}
\begin{tabular}{@{}>{\ttfamily\small\raggedright\arraybackslash}p{0.96\columnwidth}@{}}
\toprule
\normalfont\textbf{Skill Evolution Prompt} ($P_{\text{evo}}^{(i)}$, Eq.~11) \\
\midrule
You are an expert agent designer improving a skill SOP.\\[3pt]
\normalfont\textbf{Skill:} \{skill\_name\} | Success rate: \{rate:.1\%\}
(threshold: \{eta:.1\%\})\\
\normalfont\textbf{Current SOP:} \{current\_sop\}\\[3pt]
\normalfont\textbf{Critical Transition Points} (step, observations, skills,
templates, $\delta$ values): \{transition\_contexts\}\\
\normalfont\textbf{Transition Summary:} \{Delta\_i\}\\
\normalfont\textbf{Episode Outcome:} \{outcome\}\\[3pt]
\normalfont\textbf{Instructions:} Rewrite the SOP to (1)~add guidance for
situations that caused abrupt state shifts, (2)~add exit/escalation
criteria, (3)~preserve format and detail, (4)~stay within 200 words.\\[3pt]
Revised SOP: \\
\bottomrule
\end{tabular}
\caption{Skill evolution prompt. Transition diagnostics $\{\mathbf{d}_t\}$
and verbalized summary $\Delta_i$ are concatenated before the SOP rewriter.}
\label{tab:prompt_evo}
\end{table}

\subsection{Emotion Template Bank}
\label{app:templates}

Table~\ref{tab:templates} lists the $C{=}12$ natural-language emotion-state
templates and their learnable prototypes $\mathbf{q}_j$ (Eq.~3).
Templates are task-agnostic and reused across all benchmarks.

\begin{table}[ht]
\centering\small
\setlength{\tabcolsep}{4pt}
\begin{tabular}{@{}clp{4cm}@{}}
\toprule
\textbf{ID} & \textbf{State} & \textbf{Template (abbreviated)} \\
\midrule
T1  & Curious       & High curiosity; actively seeking new information. \\
T2  & Desiring      & Strong anticipation toward a specific goal. \\
T3  & Confused      & Confusion/uncertainty; current strategy may fail. \\
T4  & Approving     & Approval and confidence in the current direction. \\
T5  & Optimistic    & Positive expectation about upcoming actions. \\
T6  & Proud         & Pride and self-assurance; high internal confidence. \\
T7  & Frustrated    & Frustration/annoyance; repeated obstacles. \\
T8  & Disappointed  & Disappointment; unmet expectations. \\
T9  & Nervous       & Nervousness/apprehension; perceived risk. \\
T10 & Realizing     & Sudden realization or insight; strategy shift. \\
T11 & Amused        & Light amusement; low-stakes engagement. \\
T12 & Neutral       & Neutral baseline; no strong affective signal. \\
\bottomrule
\end{tabular}
\caption{Emotion-state template bank ($C{=}12$). The template with
highest cosine similarity to $\mathbf{r}_t$ is selected (Eq.~3) and
injected when $c_t \geq \tau$.}
\label{tab:templates}
\end{table}

\subsection{Skill Library Details}
\label{app:skill_libraries}

\paragraph{WebShop (8 skills).}
SOPs distilled from MASA training traces~\cite{yu2026masa}.
Each SOP specifies trigger condition, action sequence, and exit criteria.
\textsc{ProductSearch} (search by keywords; exit: $\geq$3 candidates),
\textsc{QueryRephrase} (reformulate when $<$2 results; exit: candidates
found or 3 attempts exhausted),
\textsc{ProductCompare} (compare attributes; exit: single preferred item),
\textsc{PriceCompare} (find cheapest compliant item),
\textsc{ReviewAnalysis} (score sentiment from reviews),
\textsc{OptionSelect} (set size/color/quantity),
\textsc{CartManage} (add to cart and verify),
\textsc{PurchaseConfirm} (confirm all constraints before purchase).

\paragraph{ALFWorld (6 skills).}
Following SkillOpt~\cite{yang2026skillopt}; one SOP per task type.
Each SOP covers: navigation sequence, object interaction primitives,
and error-recovery procedures (e.g., \textsc{Heat\,Object}: pick up
$\to$ verify appliance is open $\to$ heat $\to$ place;
\textsc{Pick\,Two}: carry objects one at a time, track first placement).

\paragraph{MATH (6 skills).}
\textsc{DirectSolve} (formula/definition; Level~1--2),
\textsc{AlgebraManip} (equation setup),
\textsc{Decompose} (multi-step sub-goals; Level~3--4),
\textsc{Casework} (exhaustive enumeration),
\textsc{GeomVisualize} (mental geometric model),
\textsc{VerifyCheck} (re-derive via alternate method).

\paragraph{MBPP (5 skills).}
\textsc{DirectImplement} (from docstring),
\textsc{DecomposeGuard} (helpers + edge-case guards),
\textsc{BruteForceOptimize} (brute-force then optimize),
\textsc{PatternRecognize} (sliding window / two-pointer / DP),
\textsc{TestDriven} (assert boundary cases first, then implement).

\section{Case Studies}
\label{app:case_studies}

We present three concrete traces illustrating how \textsc{Emotion2Skill}
changes agent behavior across skill selection, skill evolution, and
pre-action intervention.

\begin{figure*}[!t]
\centering
\begin{minipage}{0.97\textwidth}
\begin{mdframed}[linewidth=0.6pt,roundcorner=6pt,backgroundcolor=blue!2]
\small
\textbf{\normalsize Case A\enspace Emotion-Augmented Skill Selection}
\hfill\textit{WebShop\;|\;Qwen3-8B\;|\;single episode trace}

\medskip
\textbf{Scenario.}
The agent must purchase a blue cotton shirt under \$50.
Twelve results are returned initially, all priced \$38--\$78.
A na\"{i}ve price filter returns zero matches and the session risks
entering a failure loop.

\medskip
\textbf{Key observation.}
At step~3, \emph{before} the explicit ``No results'' feedback arrives,
the encoder reads T3\,Confused ($c_t{=}0.36>\tau{=}0.3$), reflecting
accumulated uncertainty about the search trajectory.
\textsc{Emotion2Skill} routes to \textsc{QueryRephrase} one step earlier
than any reactive method could; the text-only baseline repeats
\textsc{ProductSearch} and enters a failure loop at step~4.

\medskip
\begin{center}
\setlength{\tabcolsep}{5pt}\renewcommand{\arraystretch}{1.12}
\begin{tabular}{@{}c p{0.30\textwidth} p{0.15\textwidth} c c p{0.13\textwidth} p{0.13\textwidth}@{}}
\toprule
\textbf{Step} & \textbf{Observation} & \textbf{Template} & $c_t$ &
\textbf{Gate} & \textbf{Baseline} & \textbf{ES-8B} \\
\midrule
1 & Goal: blue cotton shirt $<$\$50         & T1\,Curious    & 0.62 & \checkmark & P.Search                 & P.Search \\
2 & 12 results; prices \$38--\$78           & T2\,Desiring   & 0.43 & \checkmark & P.Search                 & P.Compare \\
3 & Filter applied; 0 items $<$\$50         & T3\,Confused   & 0.36 & \checkmark & P.Search                 & \textbf{Q.Rephrase} \\
4 & ``No results for original query''       & T3\,Confused   & 0.37 & \checkmark & \textit{P.Search (loop)} & Q.Rephrase \\
5 & 6 new results; prices \$34--\$52        & T5\,Optimistic & 0.41 & \checkmark & P.Compare                & P.Compare \\
6 & Item \#3: \$38, $\bigstar$4.2, in stock & T4\,Approving  & 0.58 & \checkmark & Purchase                 & Purchase \\
\midrule
\multicolumn{5}{@{}l}{\textit{Episode outcome}} & \textbf{Failure} (loop) & \textbf{Success} \\
\bottomrule
\end{tabular}
\end{center}

\medskip
\textbf{Analysis.}
Confidence stays at $c_t \geq 0.36$ throughout, so the gate admits the signal at every step.
The Confused$\to$Confused pattern at steps~3--4
($\delta_t{=}0.42 > \bar{\delta}+\sigma_\delta$) is flagged as a transition point
and used as diagnostic evidence in Case~B.
Figure~\ref{fig:trajectory} plots the full trajectory:
(a)~$c_t$ never drops below $\tau$;
(b)~template shifts from T3 to T5 at step~5, coinciding with the successful rephrase;
(c)~$\delta_t$ spikes at step~4, marking the transition point.

\begin{center}
\includegraphics[width=0.80\textwidth]{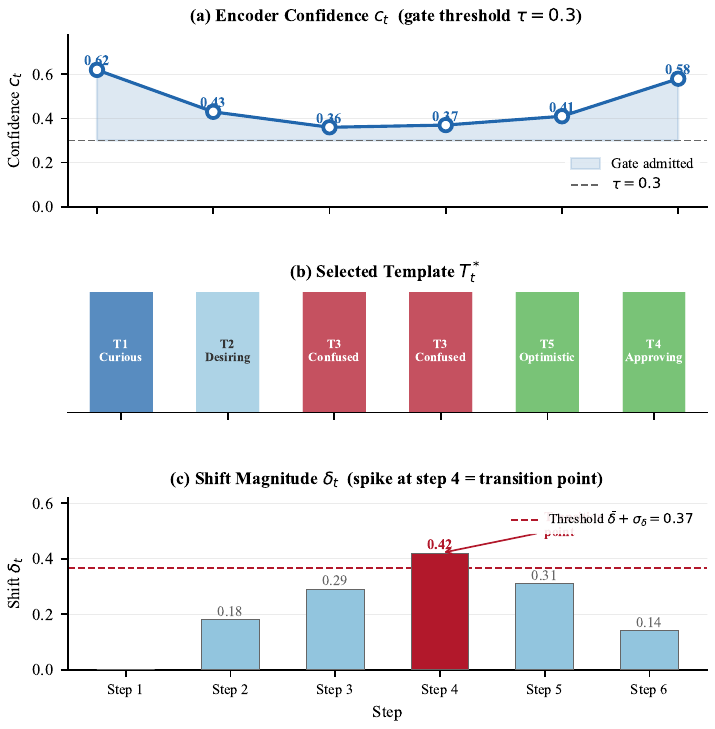}
\end{center}
\end{mdframed}
\end{minipage}
\caption{\textbf{Case A: Emotion-Augmented Skill Selection.}
The emotion encoder detects T3\,Confused at step~3 (one step before explicit failure feedback) and reroutes to \textsc{QueryRephrase}, while the text-only baseline enters a failure loop.
The bottom panel shows the full emotion trajectory including encoder confidence $c_t$, selected template $T_t^*$, and shift magnitude $\delta_t$.}
\label{app:case_selection}
\end{figure*}

\begin{figure*}[!t]
\centering
\begin{minipage}{0.97\textwidth}
\begin{mdframed}[linewidth=0.6pt,roundcorner=6pt,backgroundcolor=orange!4]
\small
\textbf{\normalsize Case B\enspace Emotion-Driven Skill Evolution}
\hfill\textit{WebShop\;|\;Qwen3-8B\;|\;Round~2 evolution}

\medskip
\textbf{Scenario.}
After round~1, \textsc{QueryRephrase} succeeds in only 38\% of episodes,
below $\eta{=}40\%$, and is flagged for evolution.
The system aggregates transition diagnostics across all 38 failed episodes
and synthesizes a verbalized summary $\Delta_i$ before rewriting the SOP.

\medskip
\textbf{Diagnostic evidence.}
Two transition points are identified in the Case~A episode:
step~4 ($\delta{=}0.42$, Confused$\to$Confused) and step~6 ($\delta{=}0.39$, Confused$\to$Confused).
Aggregating across round-1 failures, 79\% share this pattern, concentrated
at rephrasing steps with no exit condition.
The verbalized summary $\Delta_i$ produced by the LLM rewriter:

\medskip
\begin{mdframed}[linewidth=0.4pt,roundcorner=4pt,backgroundcolor=white]
\itshape\small
``The agent repeatedly enters Confused states during query rephrasing
with no exit criterion or attempt budget defined in the SOP.
Each attempt prolongs the Confused trajectory without a principled
escalation path, leading to session timeout or abandoned tasks.
The SOP should impose a maximum rephrase count and specify an explicit
fallback skill when the budget is exhausted.''
\end{mdframed}

\medskip
\textbf{SOP rewrite (Round~2).}

\medskip
\begin{center}
\setlength{\tabcolsep}{8pt}\renewcommand{\arraystretch}{1.15}
\begin{tabular}{@{}p{0.44\textwidth} p{0.50\textwidth}@{}}
\toprule
\textbf{Original SOP (Round~1)} & \textbf{Evolved SOP (Round~2)} \\
\midrule
When search results do not match the target, try synonyms or broader
terms and re-search.
Continue rephrasing until a satisfactory match is found or the task
is abandoned.
No fixed limit on rephrasing attempts.
&
\textbf{(1)}~Identify the specific failing keyword or constraint.\newline
\textbf{(2)}~Try up to \textbf{3 alternative queries}: synonyms
$\to$ broaden the core noun $\to$ drop the most restrictive attribute.\newline
\textbf{(3)}~If all 3 fail, \textbf{exit to \textsc{ProductSearch}}
with core noun and price range only.\newline
\textbf{(4)}~Never rephrase more than 3 times; escalate promptly to
prevent Confused-state loops. \\
\midrule
\multicolumn{2}{@{}p{0.96\textwidth}@{}}{\itshape
Triggers: $\delta{=}0.42$ at step~4 and $\delta{=}0.39$ at step~6
(Confused$\to$Confused in 79\% of failures).
Root cause: no rephrasing budget or exit criterion in the SOP.
Round-2 \textsc{QueryRephrase} success rate: 38\%$\to$54\% ($+$16~pp);
WebShop Succ.\ 27.2$\to$28.9\%.} \\
\bottomrule
\end{tabular}
\end{center}

\medskip
\textbf{Analysis.}
The three SOP changes (ordered query strategy, fixed attempt budget, and mandatory escalation path) are
each directly traceable to a specific finding in the diagnostic signal.
None would be discoverable from binary episode outcomes alone, which indicate only that
\textsc{QueryRephrase} failed, without localizing where or why within the episode.
\end{mdframed}
\end{minipage}
\caption{\textbf{Case B: Emotion-Driven Skill Evolution.}
Transition diagnostics identify a recurring Confused$\to$Confused pattern (79\% of failures)
and produce a verbalized summary that directly drives three targeted SOP revisions,
raising \textsc{QueryRephrase} success from 38\% to 54\%.}
\label{app:case_evolution}
\end{figure*}

\begin{figure*}[!t]
\centering
\begin{minipage}{0.97\textwidth}
\begin{mdframed}[linewidth=0.6pt,roundcorner=6pt,backgroundcolor=green!3]
\small
\textbf{\normalsize Case C\enspace Pre-Action Emotion Signal Prevents Failure}
\hfill\textit{ALFWorld Heat\;|\;Qwen3-8B\;|\;pre- vs.\ post-action routing}

\medskip
\textbf{Scenario.}
Heat tasks show the largest gain of any ALFWorld task type:
9.6\%$\to$56.9\% ($+$47.3~pp) on Qwen3-8B.
The canonical failure is issuing a heat action on a closed appliance.
We trace a representative episode: heat a mug and place it on the counter.

\medskip
\begin{center}
\setlength{\tabcolsep}{4pt}\renewcommand{\arraystretch}{1.15}
\begin{tabular}{@{}c p{0.28\textwidth} p{0.14\textwidth} c c p{0.17\textwidth} p{0.17\textwidth}@{}}
\toprule
\textbf{Step} & \textbf{Observation} & \textbf{Template} & $c_t$ &
\textbf{Gate} & \textbf{Baseline action} & \textbf{ES-8B action} \\
\midrule
1 & Task: heat mug, place on counter       & T1\,Curious    & 0.37 & \checkmark & navigate microwave          & navigate microwave \\
2 & Microwave located; door status unknown & T9\,Nervous    & 0.34 & \checkmark & \textit{heat mug} (direct)  & \textbf{examine microwave} \\
3 & Microwave is closed                    & T3\,Confused   & 0.38 & \checkmark & \textit{action failed}      & open microwave \\
4 & Microwave now open                     & T5\,Optimistic & 0.55 & \checkmark & navigate microwave (retry)  & heat mug \\
5 & Mug heated successfully                & T4\,Approving  & 0.61 & \checkmark & examine microwave           & place mug on counter \\
6 & ---                                    & ---            & ---  & ---        & \textit{loop; task fails}   & \textbf{task complete} \\
\midrule
\multicolumn{5}{@{}l}{\textit{Episode outcome}} & \textbf{Failure} & \textbf{Success} \\
\bottomrule
\end{tabular}
\end{center}

\medskip
\textbf{Key mechanism: pre-action vs.\ post-action signal.}
At step~2 the baseline issues \texttt{heat mug} immediately; failure feedback arrives
only \emph{after} the incorrect action.
\textsc{Emotion2Skill} reads T9\,Nervous ($c_t{=}0.34$), reflecting internal apprehension
about an unverified precondition, \emph{before} any action is taken, and routes to
\texttt{examine\,$\to$\,open\,$\to$\,heat}.

\medskip
\textbf{Round-1 evolution.}
The T9\,Nervous spike at step~2 ($\delta_t{=}0.38$, exceeding threshold) is flagged for
\textsc{Heat\,Object}.
Across round-1 failures, 11 of 14 failed Heat episodes share the same spike before a
failed \texttt{heat} action.
The verbalized summary $\Delta_i$:

\medskip
\begin{mdframed}[linewidth=0.4pt,roundcorner=4pt,backgroundcolor=white]
\itshape\small
``The agent consistently issues heat actions without verifying appliance state.
T9\,Nervous activations immediately before heating indicate internal uncertainty
about unresolved preconditions.
Add an explicit precondition check: verify the appliance is open before any heating action.''
\end{mdframed}

\medskip
\textbf{SOP before/after (Round~1).}

\medskip
\begin{center}
\setlength{\tabcolsep}{8pt}\renewcommand{\arraystretch}{1.15}
\begin{tabular}{@{}p{0.46\textwidth} p{0.46\textwidth}@{}}
\toprule
\textbf{Original \textsc{Heat\,Object} SOP} & \textbf{Evolved SOP (after Round~1)} \\
\midrule
Navigate to the target appliance.\newline
Pick up the object to be heated.\newline
Place the object inside the appliance.\newline
Activate heating.\newline
Retrieve the object once complete.\newline
Place the object at the goal location.
&
Navigate to the target appliance.\newline
\textbf{Verify the appliance is open; if closed, open it first.}\newline
Pick up the object to be heated.\newline
Place the object inside the appliance.\newline
Activate heating.\newline
Retrieve the object once complete.\newline
Place the object at the goal location. \\
\midrule
\multicolumn{2}{@{}p{0.94\textwidth}@{}}{\itshape
Change: one precondition check prepended (bold).
Diagnostic source: T9\,Nervous at step~2 in 11/14 failed episodes.
Impact: Heat success 9.6\%$\to$56.9\% ($+$47.3~pp).} \\
\bottomrule
\end{tabular}
\end{center}

\medskip
\textbf{Analysis.}
Outcome-only evolution (\textit{w/o Diagnostics} ablation) achieves only $+$31.4~pp on Heat,
leaving 15.9~pp attributable exclusively to the emotion diagnostic.
The signal localizes failure to a specific step within the episode and identifies it
\emph{before} the erroneous action rather than after it.
\end{mdframed}
\end{minipage}
\caption{\textbf{Case C: Pre-Action Emotion Signal Prevents Failure.}
T9\,Nervous at step~2 triggers a precondition-check reroute before the erroneous heat action.
Round-1 evolution encodes this into the SOP, raising Heat task success from 9.6\% to 56.9\%
($+$47.3~pp); 15.9~pp of this gain is attributable exclusively to emotion diagnostics
over outcome-only evolution.}
\label{app:case_alfworld}
\end{figure*}

\section{Additional Results}
\label{app:additional_results}

\subsection{Evolution Round-by-Round}
\label{app:evolution_rounds}

Table~\ref{tab:evolution_rounds} compares emotion-guided evolution
against outcome-only evolution across the three rounds.

\begin{table}[ht]
\centering\small
\setlength{\tabcolsep}{2.5pt}
\begin{tabular}{@{}p{2.5cm}ccc@{}}
\toprule
\textbf{Method} & \textbf{Round 1} & \textbf{Round 2} & \textbf{Round 3} \\
\midrule
Evo-Outcome (no diagnostics) & 26.8 & 27.5 & 27.3 \\
Evo-Emotion (\textsc{ES})    & 27.2 & 28.9 & \textbf{33.3} \\
\midrule
$\Delta$                     & $+$0.4 & $+$1.4 & $+$6.0 \\
\bottomrule
\end{tabular}
\caption{WebShop Succ.\ (\%) across evolution rounds (Qwen3-8B).
Emotion diagnostics yield compounding gains; the gap widens to
$+$1.6\,pp by round~3 as more transition evidence accumulates.}
\label{tab:evolution_rounds}
\end{table}

\subsection{Per-Task Emotion Patterns on ALFWorld}
\label{app:per_task_emotions}

Table~\ref{tab:per_task_emotions} shows the most frequently activated
template per ALFWorld task type on correctly-solved episodes.
Procedure-heavy tasks (Heat, PickTwo) activate frustration/confusion,
consistent with their multi-step error-recovery demands.

\begin{table}[ht]
\centering\small
\begin{tabular}{@{}lcc@{}}
\toprule
\textbf{Task} & \textbf{Top Template} & \textbf{Freq.} \\
\midrule
Pick    & T1 Curious    & 42\% \\
Look    & T4 Approving  & 38\% \\
Clean   & T3 Confused   & 35\% \\
Heat    & T7 Frustrated & 41\% \\
Cool    & T10 Realizing & 33\% \\
PickTwo & T3 Confused   & 39\% \\
\bottomrule
\end{tabular}
\caption{Most frequent emotion template per ALFWorld task type
(Qwen3-8B, correctly-solved episodes only).}
\label{tab:per_task_emotions}
\end{table}

\subsection{Confidence Score Distribution}
\label{app:confidence_dist}

\begin{figure}[ht]
\centering
\includegraphics[width=\columnwidth]{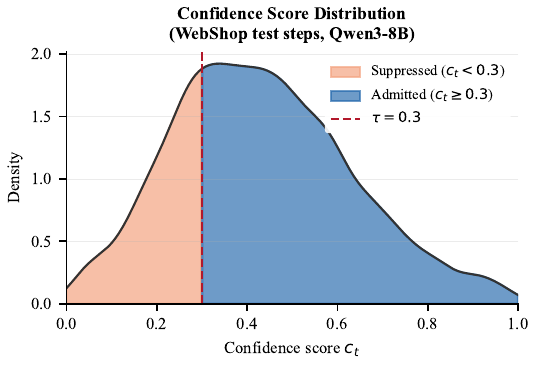}
\caption{Confidence score distribution (WebShop, Qwen3-8B).
Mass concentrated in $c_t \in [0.30, 0.60]$ with a sparse high tail;
72\% of steps are gate-admitted (blue), 28\% suppressed (orange).
Gate threshold $\tau{=}0.3$ shown as dashed line.}
\label{fig:confidence_hist}
\end{figure}

Figure~\ref{fig:confidence_hist} shows the distribution of encoder
confidence scores $c_t$ across all WebShop test steps.
The distribution is right-skewed, with the bulk of steps concentrated
in the range $c_t \in [0.30, 0.60]$ and a sparse high-confidence tail
above 0.70.
A substantial minority of steps (28\%) fall below $\tau{=}0.3$ and
trigger the text-only fallback, reflecting genuine uncertainty at
ambiguous or transitional states.
The remaining 72\% are admitted by the gate and benefit from
emotion-augmented skill selection.

\section{Limitations and Broader Impact}
\label{app:limitations}

\paragraph{Limitations.}
White-box residual-stream access is required, limiting direct application
to API-served LLMs; distilling routing signals to black-box models is
a natural extension.
The warm-up phase collects $\sim$500 episodes ($<$10\,min); few-shot
encoder initialization could reduce this.
Emotion-guided routing does not uniformly dominate all task types:
on ALFWorld Cool tasks, \textsc{Emotion2Skill} is outperformed by MASA
on Qwen3-8B, suggesting that tasks with highly stereotyped action sequences
may benefit less from emotion-based rerouting and could benefit from
adaptive per-task thresholding as future work.
Instruction-tuned variants with modified hidden-state geometry may
require re-extraction (Qwen3-8B-Instruct GoEmotions accuracy: 28.1\%
vs.\ 37.2\% on the base model).
Extension to multi-agent coordination is left open.

\paragraph{Broader Impact.}
Making LLM internal states actionable improves interpretability and
debuggability of agent systems.
In high-stakes domains (healthcare, finance), the confidence-gating
mechanism provides a built-in safety fallback when the emotion signal
is uncertain.
Access to internal model states could theoretically be misused for
surveillance; responsible deployment requires appropriate access controls.
On balance, we believe the transparency and adaptive-control benefits
outweigh the risks.

\section{Notation Summary}
\label{app:notation}

Table~\ref{tab:notation} provides a consolidated reference for all
mathematical symbols used in the paper, grouped by subsystem.

\begin{table*}[ht]
\centering\small
\setlength{\tabcolsep}{5pt}
\renewcommand{\arraystretch}{1.12}
\begin{tabular}{@{}lp{9.8cm}@{}}
\toprule
\textbf{Symbol} & \textbf{Definition} \\
\midrule
\multicolumn{2}{@{}l@{}}{\textit{Agent and task}} \\
$\mathcal{M}$ & LLM agent backbone (Qwen3-8B or Qwen3-14B, non-thinking mode) \\
$\mathcal{S} = \{s_1,\ldots,s_M\}$ & Skill library of $M$ skills; $D_i$: natural-language SOP for skill $s_i$ \\
$x_t$ & External textual context at step $t$ (task instruction, history, observation) \\
$s_t^*$ & Skill selected at step $t$ \\
$r_{\text{ep}} \in \{0,1\}$ & Episode outcome (1 = success) \\
\midrule
\multicolumn{2}{@{}l@{}}{\textit{LLM internals}} \\
$\mathbf{h}_t^{(\ell)} \in \mathbb{R}^d$ & Residual-stream activation at layer $\ell$, token position $t$; $d$: hidden dim \\
$L^*$ & Optimal extraction layer (selected by GoEmotions validation accuracy) \\
$\bar{\mathbf{h}}^{(\ell)}$ & Mean-pooled per-story activation at layer $\ell$ (tokens $t\!\geq\!50$) \\
\midrule
\multicolumn{2}{@{}l@{}}{\textit{Emotion direction extraction (offline)}} \\
$K = 27$ & Number of GoEmotions emotion concepts \\
$N = 100$ & Stories generated per emotion concept \\
$\mu_k^{(\ell)}$ & Within-class mean activation for concept $k$ at layer $\ell$ \\
$\bar{\mu}^{(\ell)}$ & Global mean over all $KN$ story activations at layer $\ell$ \\
$\mathbf{v}_k^{(\ell)} \in \mathbb{R}^d$ & Contrastive emotion direction: $\mu_k^{(\ell)} - \bar{\mu}^{(\ell)}$ (Eq.~1) \\
$\Pi^\perp$ & PCA denoising projector (removes top neutral-activation components) \\
$\hat{\mathbf{v}}_k^{(\ell)}$ & Denoised direction: $\Pi^\perp \mathbf{v}_k^{(\ell)}$ \\
$W_e \in \mathbb{R}^{K \times d}$ & Emotion extractor matrix; rows $= \hat{\mathbf{v}}_k^{(L^*)}$ \\
\midrule
\multicolumn{2}{@{}l@{}}{\textit{Online emotion state}} \\
$\mathbf{e}_t \in \mathbb{R}^K$ & $\ell_2$-normalized emotion state at step $t$: $W_e\mathbf{h}_T^{(L^*)}/\|W_e\mathbf{h}_T^{(L^*)}\|_2$ (Eq.~2) \\
$f_\theta$ & Emotion encoder: 3-layer MLP ($27{\to}64{\to}128{\to}128$, ReLU, $<$27K params) \\
$\mathbf{r}_t \in \mathbb{R}^{128}$ & Encoded representation: $f_\theta(\mathbf{e}_t)$ \\
\midrule
\multicolumn{2}{@{}l@{}}{\textit{Template selection and confidence gating}} \\
$C = 12$ & Number of emotion-state templates \\
$\mathcal{T} = \{T_1,\ldots,T_C\}$ & Template bank (task-agnostic; Table~\ref{tab:templates}) \\
$\mathbf{q}_j \in \mathbb{R}^{128}$ & Learnable prototype for template $T_j$ \\
$T_t^*$ & Selected template: $\arg\max_j \cos(\mathbf{r}_t, \mathbf{q}_j)$ (Eq.~3) \\
$c_t \in [0,1]$ & Confidence score: $\sigma(\mathbf{w}_c^\top \mathbf{r}_t + b_c)$ (Eq.~4) \\
$\tau$ & Confidence gate threshold (default 0.3, tuned on WebShop validation) \\
$\boldsymbol{\phi}_t = (T_t^*, c_t)$ & Composite emotion summary injected into $P_t$ \\
$P_t$ & Augmented prompt at step $t$ (Eq.~5); reduces to text-only when $c_t < \tau$ \\
\midrule
\multicolumn{2}{@{}l@{}}{\textit{Encoder training losses}} \\
$\mathcal{A}^+(i)$ & Positive set for anchor $i$: same skill \& successful outcome \\
$\mathcal{L}_{\text{ctr}}$ & Supervised-contrastive loss (Eq.~6); temperature $\kappa{=}0.07$ \\
$\mathcal{L}_{\text{cls}}$ & Template-classification loss (Eq.~7); temperature $\kappa'{=}0.1$ \\
$\lambda$ & Loss weight: $\mathcal{L} = \mathcal{L}_{\text{ctr}} + \lambda\mathcal{L}_{\text{cls}}$, default $\lambda{=}0.5$ \\
\midrule
\multicolumn{2}{@{}l@{}}{\textit{Skill evolution}} \\
$\mathcal{E} = (\mathbf{e}_1,\ldots,\mathbf{e}_T)$ & Episode emotion trajectory \\
$\delta_t = 1 - \mathbf{e}_t^\top\mathbf{e}_{t-1}$ & Step-to-step cosine distance (Eq.~8) \\
$\bar{\delta},\,\sigma_\delta$ & Episode mean and std of $\delta_t$ values \\
$\mathcal{P}$ & Transition-point set: $\{t \mid \delta_t > \bar{\delta}+\sigma_\delta\}$ (Eq.~9) \\
$\mathcal{P}_i$ & Skill-specific transitions: $\{t \in \mathcal{P} \mid s_t^* = s_i\}$ \\
$w$ & Diagnostic window half-width (default $w{=}1$; 3-step context) \\
$\mathbf{d}_t$ & Diagnostic context: $(x_{t-w:t+w},\,s_{t-w:t+w}^*,\,T_{t-w:t+w}^*,\,\delta_{t-w:t+w})$ (Eq.~10) \\
$\Delta_i$ & Verbalized transition summary for skill $s_i$ \\
$P_{\text{evo}}^{(i)}$ & Evolution prompt for skill $s_i$ (Eq.~11) \\
$D_i'$ & Revised SOP generated by the LLM rewriter (Eq.~12) \\
$\mathcal{U}^{(r)}$ & Underperforming skills in round $r$ (success rate $< \eta$) \\
$\eta$ & Performance threshold for flagging skills (default 0.4) \\
$R$ & Number of evolution rounds (default 3) \\
\bottomrule
\end{tabular}
\caption{Complete notation reference.
Symbols are grouped by the pipeline stage in which they first appear.
Standard notation ($\mathcal{M}$, $D_i$, $x_t$) is also defined inline where first used in the main text.}
\label{tab:notation}
\end{table*}

\end{document}